\documentclass[journal]{IEEEtran}
\usepackage{amsmath,graphicx}
\usepackage{epsfig}
\usepackage{amsfonts}
\usepackage{xcolor}
\usepackage{soul}
\usepackage[utf8]{inputenc}
\usepackage[small]{caption}
\usepackage{mathtools}
\usepackage{amsfonts}
\usepackage{epstopdf}
\usepackage{makecell}
\usepackage{subfig}
\usepackage{diagbox}
\usepackage{multirow}
\usepackage{multicol}

\newcommand{\PreserveBackslash}[1]{\let\temp=\\#1\let\\=\temp}

\newcolumntype{C}[1]{>{\PreserveBackslash\centering}p{#1}}
\newcolumntype{R}[1]{>{\PreserveBackslash\raggedleft}p{#1}}
\newcolumntype{L}[1]{>{\PreserveBackslash\raggedright}p{#1}}

\usepackage[pagebackref=false,breaklinks=true,letterpaper=true,colorlinks,bookmarks=false,citecolor=blue]{hyperref}

\ifCLASSINFOpdf
\else
\fi

\newcommand{\HF}[1]{\textcolor[rgb]{0.00,0.00,0.00}{#1}}

\begin{document}
%\title{Bare Demo of IEEEtran.cls\\ for IEEE Journals}
\title{Robust and Efficient Graph Correspondence Transfer for Person Re-identification}

\author{Qin~Zhou,
        Heng~Fan,
        Hua~Yang,~\IEEEmembership{Member,~IEEE,}
          Hang~Su,~\IEEEmembership{Member,~IEEE,}
        Shibao~Zheng,~\IEEEmembership{Member,~IEEE,}
        Shuang~Wu,
        and~Haibin~Ling,~\IEEEmembership{Member,~IEEE}% <-this % stops a space
\thanks{Qin Zhou, Shibao Zheng and Hua Yang are with Institute of Image Processing and Network Engineering, Shanghai Jiao Tong University, Shanghai 200240, China. Email: \{zhou.qin.190, sbzh, hyang\}@sjtu.edu.cn.}% <-this % stops a space
\thanks{Heng Fan and Haibin Ling are with Department of Computer \& Information Sciences, Temple University, Philadelphia 19122, USA. Email: \{hengfan, hbling\}@temple.edu.}
\thanks{Hang Su is with Tsinghua University, Beijing 100084, China. Email: suhangss@mail.tsinghua.edu.cn.}
\thanks{Shuang Wu is with YouTu Lab, Tencent, Shanghai 200233, China. Email: calvinwu@tencent.com.}
\thanks{
Shibao Zheng and Hua Yang are the corresponding authors.}
% \thanks{This work was done when Qin Zhou was a visiting student at Temple University in Professor Haibin Ling's group.}

}


\markboth{Draft}%
{Shell \MakeLowercase{\textit{et al.}}: Bare Demo of IEEEtran.cls for IEEE Journals}



\maketitle

\begin{abstract}
 Spatial misalignment caused by variations in poses and viewpoints is one of the most critical issues that hinders the performance improvement in existing person re-identification (Re-ID) algorithms. To address this problem, in this paper, we present a robust and efficient graph correspondence transfer (REGCT) approach for explicit spatial alignment in Re-ID. Specifically, we propose to establish the patch-wise correspondences of positive training pairs via graph matching. By exploiting both spatial and visual contexts of human appearance in graph matching, meaningful semantic correspondences can be obtained. To circumvent the cumbersome \emph{on-line} graph matching in testing phase, we propose to transfer the \emph{off-line} learned patch-wise correspondences from the positive training pairs to test pairs. In detail, for each test pair, the training pairs with similar pose-pair configurations are selected as references. The matching patterns (i.e., the correspondences) of the selected references are then utilized to calculate the patch-wise feature distances of this test pair. To enhance the robustness of correspondence transfer, we design a novel pose context descriptor to accurately model human body configurations, and present an approach to measure the similarity between a pair of pose context descriptors. Meanwhile, to improve testing efficiency, we propose a correspondence template ensemble method using the voting mechanism, which significantly reduces the amount of patch-wise matchings involved in distance calculation. With aforementioned strategies, the REGCT model can effectively and efficiently handle the spatial misalignment problem in Re-ID. Extensive experiments on five challenging benchmarks, including VIPeR, Road, PRID450S, 3DPES and CUHK01, evidence the superior performance of REGCT over other state-of-the-art approaches.

 %  With such a strategy, the proposed REGCT model can effectively handle the spatial misalignment problem in an efficient way.
\end{abstract}


% \fi
\IEEEpeerreviewmaketitle
\section{Introduction}
\IEEEPARstart{P}{erson} re-identification (Re-ID), which aims to associate a probe image to images in the gallery set (usually across different non-overlapping camera views), plays a crucial role in various applications including video surveillance, human-computer interaction, etc. Despite great successes in recent years, accurate Re-ID remains challenging due to many factors such as large human appearance changes in different camera views and heavy body occlusions. To deal with these issues, numerous Re-ID approaches have been proposed~\cite{FarenzenaBPMC10,KaranamLR15,salicency,sim_spatial_constraints,KISSME,SVMML,lomo,pcca,learn_to_rank,lfda,TMA}.

\begin{figure}[!t]
  \centering
  \includegraphics[width=\linewidth]{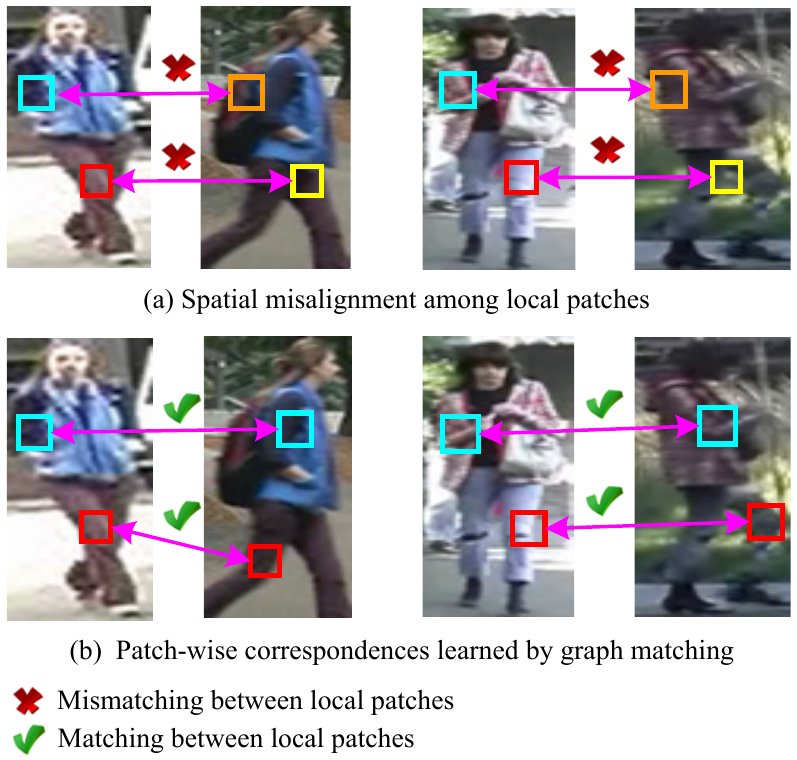}
    % \captionsetup{font={footnotesize}}
  \caption{Illustration of the spatial misalignment problem in Re-ID. Image (a) illustrate the spatial misalignment problem (i.e., spatially corresponding patches do not indicate correct semantic patch-wise matching) caused by pose and viewpoint changes. The proposed REGCT model can capture the correct semantic matching among patches using patch-wise graph matching, as shown in image (b).}
  \label{GCT_insight}
\end{figure}

For Re-ID, a major challenge is to deal with the inevitable spatial misalignment problem between image pairs caused by large variations in camera views and human poses, as shown in Fig.~\ref{GCT_insight}. Most existing methods~\cite{YangYYLYL14,kLFDA14,ChenYHZW15}, nevertheless, focus on addressing the problem of Re-ID by comparing the holistic visual differences between images, which ignore the spatial misalignment problem. To alleviate this issue, there are some attempts to apply part-based approaches to handle misalignment~\cite{iccv15_correspondence,OreifejMS10,salicency,YangWLL17}. These methods divide objects into local patches and perform an {\it on-line} patch-level matching for Re-ID. Though these approaches can handle spatial misalignment to some extent, being in lack of modeling the spatial and visual context information among local patches during correspondence learning, they still fail in presence of visually similar body appearances, or occlusions. Some other algorithms divide the images into fixed stripes, and extract visual statistics from each stripe, assuming that the human body is centered and consistently cropped within the bounding boxes~\cite{lomo,sim_spatial_constraints}. However, it is often not the case that human body is optimally cropped, especially when the bounding boxes are generated by pedestrian detection algorithms.

%To address the above-mentioned issues, we propose to alleviate the spatial misalignment problem via patch-wise graph matching. We adopt the local patches as matching units in that patches are middle-level regions that can both capture informative visual appearances and still have enough flexibility for establishing correspondences. Since it is meaningless to perform graph matching between patches of negative image pairs, we propose a novel framework of{\it off-line} patch-level matching templates for positive training pairs via graph matching,
\begin{figure}[!t]
  \centering
  \includegraphics[width=0.8\linewidth]{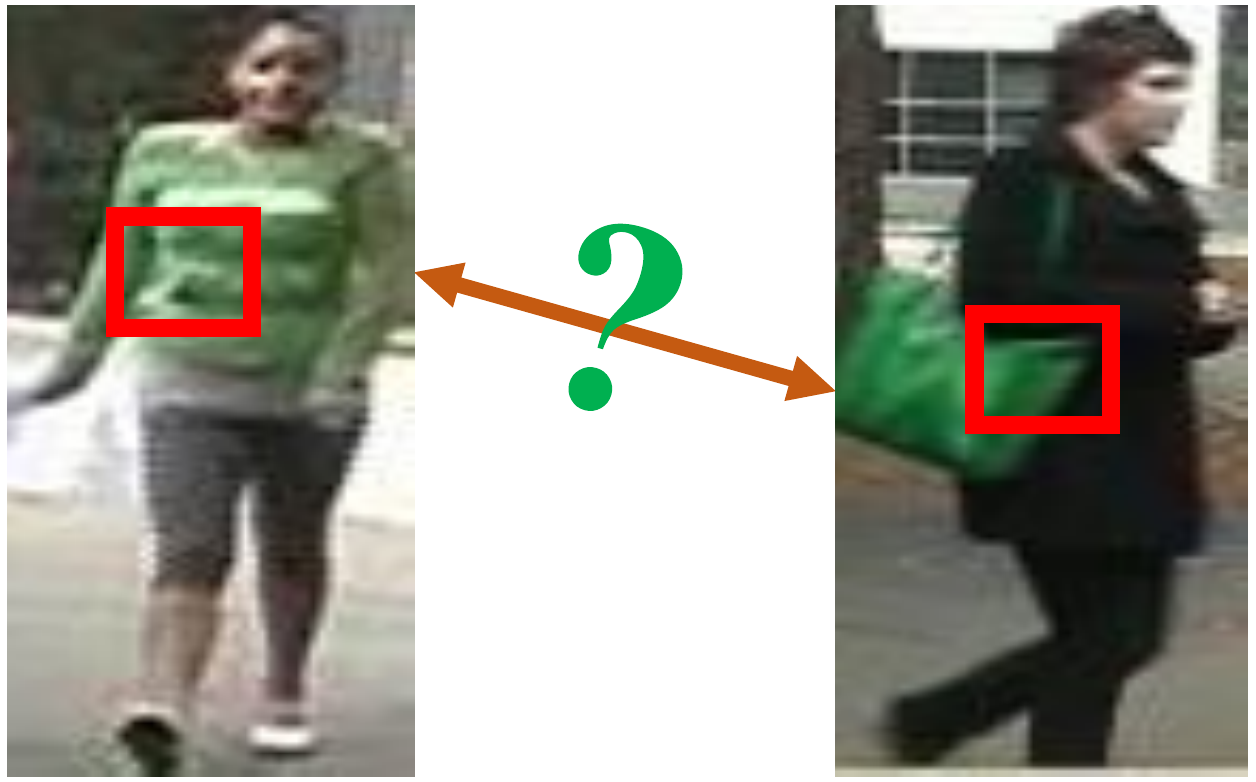}
   \captionsetup{font={footnotesize}}
  \caption{Illustration of the fact that \emph{on-line} graph matching is not suitable for generating semantic correspondences between negative pairs.}
  \label{fig:reason_transfer}
%  \vspace{-1em}
\end{figure}

Intuitively, patch-wise correspondences are spatially and visually compatible to make reasonable semantic matchings. As shown in Fig.~\ref{GCT_insight}(b), spatial compatibility means that the semantically corresponding patch is supposed to exist in the local neighborhood of the probe patch in the image plane (This ensures that a patch of the head should not be matched with a patch on the leg). On the other hand, visual compatibility indicates that semantic matchings should be visually similar. Based on these two basic assumptions, \HF{we propose to  automatically discover the patch-level semantic matching patterns for each positive training pair via graph matching}. In our model, both spatial and visual contexts are taken into consideration to establish accurate semantic patch-wise matching results.  By using the part-based strategy and implicitly modeling body context information into graph matching, our REGCT algorithm is able to deal with the spatial misalignment problem.

Although graph matching is straightforward for establishing semantic correspondences between positive image pairs, we argue that it is not suitable for negative ones. As shown in Fig.~\ref{fig:reason_transfer}, \emph{on-line} graph matching may establish correspondences between the two red patches, since they are spatially and visually compatible within this image pair. However, they are not semantic matchings (one belongs to the torso, while the other belongs to the bag). To address this issue, instead of directly estimating the patch-wise matchings between each test pair, we propose to transfer the \emph{off-line} learned patch-wise matchings from the positive training pairs to the test pairs for performance evaluation.

\begin{figure}[!t]
	\centering
	\includegraphics[width=\linewidth]{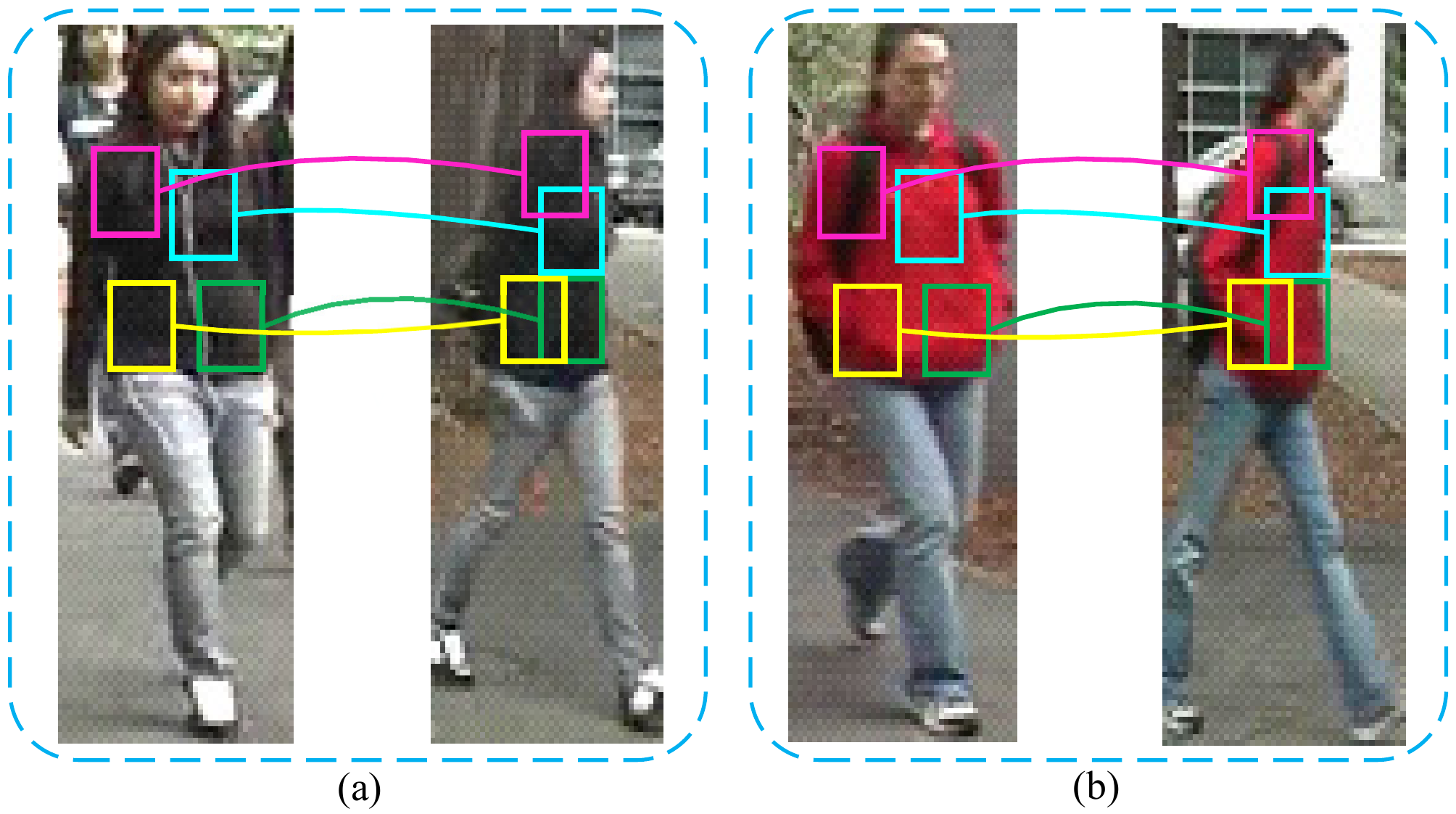}
	\captionsetup{font={footnotesize}}
	\caption{The observation that two image pairs with similar pose-pair configurations tend to share similar patch-wise correspondences. For example, the left/right images in (a) have similar poses w.r.t. the left/right image in (b). Therefore, the patch-wise matchings denoted by bounding boxes of the same colors in (a) are similar to matchings in (b). Best viewed in color.}
	\label{fig:observation}
	%   \vspace{-1em}
\end{figure}

The concept of correspondence transfer is based on the observation that two image pairs with similar pose-pair configurations tend to share similar patch-level correspondences (as shown in Fig.~\ref{fig:observation}). To better represent the human body configurations, we design a novel pose context descriptor to capture the spatial context of body joints. For each pair of test samples, their pose context descriptors are compared with the pose context descriptors of all the positive training pairs, and the training pairs with the most similar pose-pair configurations are selected as references. Finally, the matching patterns of those referred training pairs are utilized to compute the overall feature distance between this test pair.

%The patch-wise Mahalanobis feature distance calculation is time-consuming. Given that multiple training pairs are selected as references and all their patch-wise correspondence patterns are counted in the final distance calculation. Therefore, the evaluation time grows linearly as the number of the reference increases. Meanwhile,~\cite{our_gct/Zhou18} shows in the experimental part that generally more reference pairs lead to better performance. Therefore, to reduce the computational load as well as to improve the robustness of the matching result, we present a voting based matching template ensemble approach to aggregate the original noisy patch-wise correspondences into a more compact result.

In summary, we make the following contributions: (1) We for the first time propose a novel robust and efficient graph correspondence transfer (REGCT) model for Re-ID, which takes into account both spatial and visual contexts to handle the spatial misalignment problem by establishing semantic patch-wise matchings between positive training pairs.
%Spatial and visual context information are considered to explore the semantic patch-wise correspondences between positive image pairs to handle the spatial misalignment problem. And for the first time, a novel graph correspondence transfer (GCT) model is presented for person re-identification.
%\item A novel intra-pair and inter-pair matching consistency constraint is introduced for post-matching refinement to further improve the accuracy of graph matching.
(2) We introduce the pose context descriptor to accurately model the body configurations for more robust correspondence transfer.
%Based on the observation that image pairs with similar pose-pair configurations tend to share similar patch-level correspondences,
(3) We present a voting based strategy to integrate multiple noisy correspondence templates into a more compact patch-wise matching pattern, which not only reduces the computational load, but also improves the robustness of correspondence transfer.
(4) Extensive experiments on five benchmarks demonstrate that our REGCT model performs favorably against state-of-the-art approaches, and in fact even better than many deep learning based solutions.

This paper is an extended version of our preliminary work~\cite{our_gct/Zhou18}. The main differences from~\cite{our_gct/Zhou18} include: (1) For correspondence transfer, we propose a novel pose context descriptor based on the topology structure of the estimated joint locations~\cite{real_time_pose_estimation/CaoSWS17}, which improves the robustness of correspondence transfer and demonstrates superior recognition performance compared with the body orientation based transfer in~\cite{our_gct/Zhou18}. (2) We present a voting based approach to integrate multiple noisy correspondence templates into a compact matching pattern, resulting in computation reduction as well as robustness improvement during testing. (3) We conduct more ablative studies to analyze each component in our REGCT model and give insights into the best configurations of different parameter settings. (4) Notable performance gains are obtained with the above-mentioned contributions compared to~\cite{our_gct/Zhou18}.

\section{Related Work}
% \subsection{Person Re-identification}
Being extensively studied, Re-ID has drawn extensive attention in the past years. For a comprehensive survey, please refer to~\cite{survey/ZhengYH16,survey/Bedagkar-GalaS14}. In this section, we briefly review existing Re-ID algorithms from three perspectives: (1) feature representation based methods that focus on designing sophisticated features to better represent the human appearance, (2) metric learning based algorithms that pursue discriminative subspaces where features of the same person sit closer than those of different individuals, and (3) deep learning based approaches that aim to learn discriminative representations through end-to-end deep architecture modeling.

\begin{figure*}[!t]
  \centering
  \includegraphics[width=\linewidth]{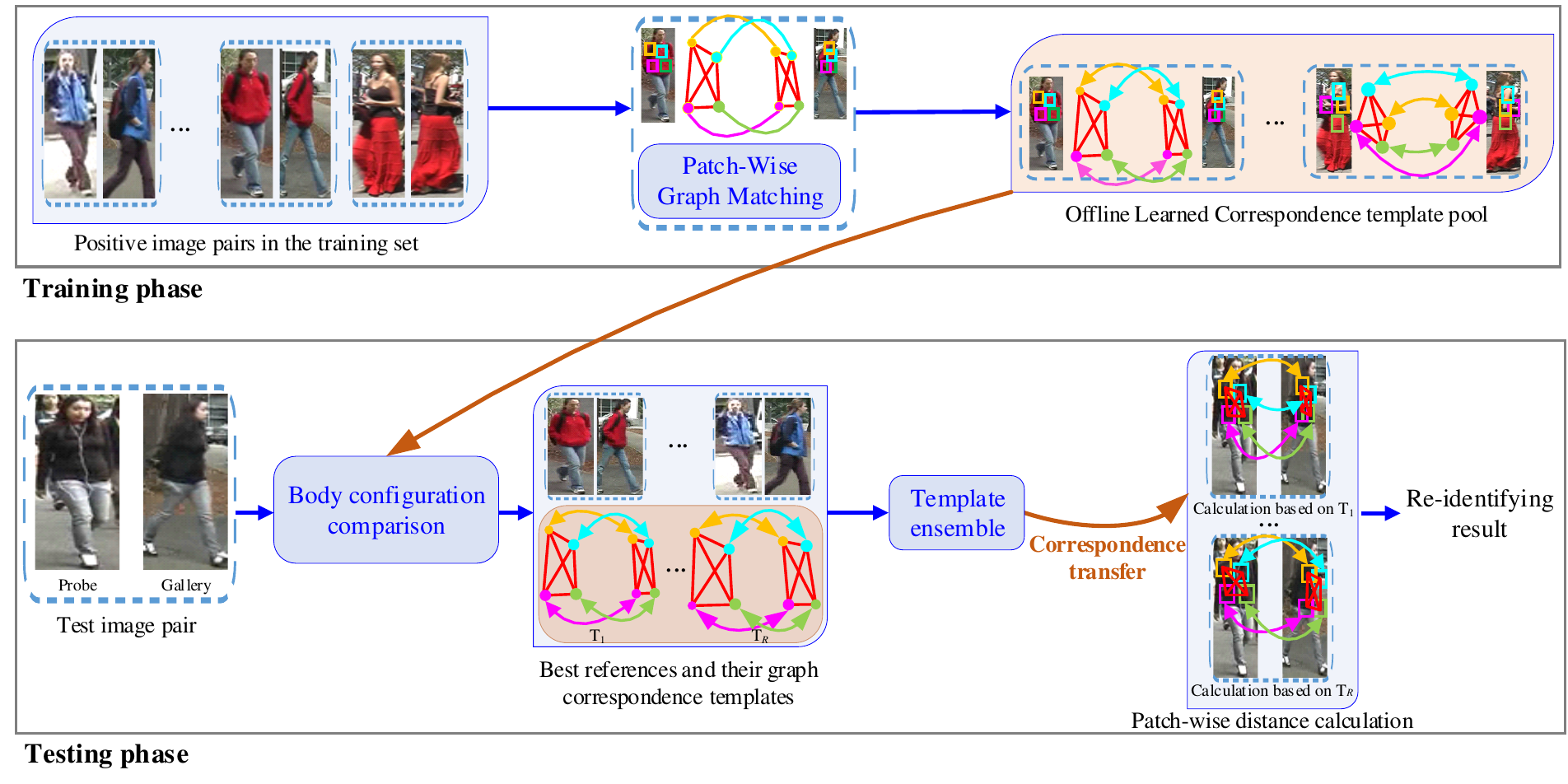}
%\captionsetup{font={footnotesize}}
%\vspace{-1ex}
  \caption{Illustration of the REGCT model. During training, spatial and visual context information are embedded into graph matching to establish patch-wise matchings for positive training pairs with various pose-pair configurations. %To improve the matching accuracy, intra-pair and inter-pair matching consistency constraint is introduced for correspondence refinement.
  During testing, for a pair of test samples, we choose a few positive training pairs with the most similar pose-pair configurations as references, and then transfer the correspondences of these references to this test pair for feature distance calculation. \textbf{Different from the preliminary work~\cite{our_gct/Zhou18}, a novel pose context descriptor is proposed for more accurate body configuration
comparison. Besides, the templates
ensemble method is introduced to pursue robust and efficient correspondence transfer during testing.}}\label{GCT_model}
% \vspace{-2ex}
\end{figure*}

\subsubsection{Feature Designing Algorithms}
The early works on person re-identification focus on designing representative features to pursue identity invariance across different cameras. In order to improve the representative and discriminative ability of hand-crafted features, various visual cues are exploited. In~\cite{FarenzenaBPMC10}, three complementary visual cues are used to model the human appearance: the overall chromatic content, the spatial arrangement of colors into stable regions, and the presence of recurrent local motifs with high entropy. Besides, symmetry and asymmetry body structure information are applied to localize the vertical axis of human body, and the local features are reweighted by the distance with respect to the
vertical axis such that the effects of pose variations are minimized. The work of~\cite{spatial_temporal_feature/GheissariSH06} proposes a spatio-temporal segmentation algorithm to generate salient edges and combines these salient edges with normalized colors to design invariant signatures. Other feature descriptors, including fisher vector encoded local descriptor~\cite{LDFV/MaSJ12}, HPEsignature~\cite{hpe/BazzaniCPFM10} and mean riemannian
covariance grid~\cite{riemannian/BakCBT11}, are also introduced to better represent the human appearance. Even though carefully designed, hand-crafted features are limited in modeling human appearance in complicated scenes. Therefore, some researchers resort to leveraging learning techniques to model middle or high-level information of human appearance. Representative works include attribute assisted clothes appearance~\cite{attribute/LiLY14}, dictionary learning base features~\cite{dictionary/LiuSTZCB14,dictionary/KaranamLR15} etc.

\subsubsection{Metric Learning Algorithms}
Metric learning methods, on the other hand, aim to learn an optimal subspace where the intra-person divergence is minimized and meanwhile the inter-person divergence is maximized. In~\cite{KISSME}, a simple while effective strategy is proposed to infer the distance from equivalence constraints. In requirement of no iterations for optimization, the algorithm of~\cite{KISSME} runs efficiently and can benefit large scale re-identification. PCCA~\cite{pcca} presents a method to learn a low-dimensional discriminative subspace from sparse pairwise similar/disimilar constraints. The authors further introduce the ``kernel trick'' to generalize PCCA to the nonlinear cases. Local fisher discriminant analysis is introduced in~\cite{lfda} to perform discriminative feature dimension reduction based on which the intra-class instances are pulled together while inter-class ones are pushed apart. To better leverage the advantages of different metric learning methods, the work of~\cite{learn_to_rank} seeks metric ensembles by learning to rank techniques. To alleviate the spatial misalignment problem,~\cite{sim_spatial_constraints} proposes to learn separate sub-similarity functions for different sub-regions with the help of polynomial feature maps, and complementary strength of local similarities as well as global similarity are combined together for better matching consistency.

\subsubsection{Deep Learning for Re-ID}
Recently deep convolutional features have been demonstrated to significantly boost the performance of various computer vision tasks including object detection~\cite{detection/HeZRS16,faster_rcnn/RenHGS15}, tracking~\cite{fan2017parallel,fan2017sanet}, object segmentation~\cite{segmentation/LiuLLLT15,fan2018multi}, etc. Inspired by the powerful ability of deep features, many researchers resort to building deep end-to-end architectures to directly learn discriminative high-level features for Re-ID. In~\cite{deep_reid/LiZXW14}, the authors propose a novel filter pairing neural network (FPNN) to jointly handle misalignment, photometric and geometric transforms, occlusions and background clutter by designing corresponding layers to take charge of each aspect. Ahmed et al.~\cite{AhmedJM15} introduce a novel layer that computes cross-input neighborhood differences to capture local relationships between the two input images, and the cross-input neighborhood differences are aggregated together to form a pairwise cross-view representation for a pair of inputs. The work of~\cite{deep_ranking} introduces a unified deep learning-to-rank framework that learns joint representation and similarities of image pairs directly from image pixels. In~\cite{XiaoLOW16}, Xiao et al. present a novel domain guided dropout algorithm to learn robust feature representations by leveraging information from multiple domains.
%Semantic parts~\cite{deep_aligned/ZhaoLZW17,LiC0H17} or pose~\cite{pose_driven/SuLZX0T17} based deep learning algorithms are also proposed to explicitly tackle the spatial misalignment problem.

The proposed algorithm belongs to the non-deep-learning group. Instead of designing sophisticated features or pursing discriminative distance metrics, we present a novel framework to first establish local patch-wise correspondences, then aggregate the patch-wise feature similarities for recognition. This framework enjoys the flexibility that most of the aforementioned algorithms can be incorporated as part of it to improve the final recognition performance. More specifically, off-the-shelf hand-crafted or deep features can be used as visual cues for graph nodes (image patches in our case) to build the affinity matrix for graph matching. Once the local correspondences are established, existing metric learning algorithms can be adopted to pursue better similarity functions for calculating patch-wise matching scores.

The most relevant work to ours is~\cite{iccv15_correspondence} in which a correspondence structure learning (CSL) method is proposed for Re-ID. However, our REGCT model significantly differs from CSL~\cite{iccv15_correspondence} in two aspects: (i) Instead of learning a holistic correspondence structure for each camera pair in CSL~\cite{iccv15_correspondence}, we leverage graph matching to establish accurate patch-wise correspondences for each positive image pair in the training stage, and then transfer the learned matching patterns for distance calculation during testing. (ii) We model the spatial and visual context information in the affinity matrix for graph matching, which is neglected in CSL~\cite{iccv15_correspondence}. Due to the flexible and accurate instance-specific patch-wise correspondence learning and transfer, our algorithm demonstrates superior performance over CSL~\cite{iccv15_correspondence} on all the five benchmarks.

\section{Our Approach}
In this section, we first give a brief introduction on graph matching, and then elaborate on the proposed REGCT algorithm, which is composed of correspondence learning, reference selection, patch-wise feature distance calculation and aggregation based on correspondence transfer. The overall framework of the proposed REGCT algorithm is illustrated in Fig.~\ref{GCT_model}.
 \vspace{-2ex}
\subsection{Graph Matching}
Graph matching is a fundamental problem closely related to many computer vision tasks including feature registration~\cite{gm_correspondence/TorresaniKR08}, object recognition~\cite{gm_recognition/DuchenneBKP11}, visual tracking~\cite{graph_matching_tracking/tao} and so on. In this paper, we omit the detailed literature on graph matching, and only present the commonly adopted formulation to provide some insights into how graph matching is utilized to establish correspondences. For a detailed survey on graph matching, please refer to~\cite{gm_survey/ZhouT16b,gm_survey/TaoW}.

Generally, a graph \small $G=(V, E)$ of size $n$ is defined on a finite set of $n$ vertices \small $V = \{v_i\}_{i=1}^{n}$ and edges \small $E\subset V \times V$. %For each $G$, a corresponding symmetric adjacency matrix $Z\in R^{n\times n}$ is equivalently defined to capture the structure of the graph, where $Z_{ij} =1$ if there exists an edge between $v_i,v_j$, and $Z_{ij} =0$ otherwise.
For two graphs \small $G_1=(V_1, E_1)$ of size $n_1$ and \small $G_2=(V_2, E_2)$ of size $n_2$, graph matching aims to find an optimal assignment matrix \small $X \in \{0,1\}^{n_1 \times n_2}$, where $X_{ij}=1$ indicates an established correspondence between node $i$ in $G_1$ and node $j$ in $G_2$. $X$ can be optimized by maximizing the following objective:

\begin{equation}\small
\arg \mathrm{\max\limits_\mathbf{x}} {\kern 1pt}{\kern 1pt}{\kern 1pt}{\kern 1pt}  {\kern 1pt}{\kern 1pt} \mathbf{x}^{\mathrm{T}}K\mathbf{x},
\label{eq:concise_formulation}
\end{equation}
where $\mathbf{x} \in \{0,1\}^{n_1n_2}$ is the vector form of the assignment matrix $X$, and \small $K\in R^{n_1n_2\times n_1n_2}$ represents the affinity matrix that encodes both the node similarity and edge compatibility information between $G_1$ and $G_2$.
 \vspace{-2ex}
\subsection{Patch-wise correspondence learning with graph matching}
In this paper, we adopt the attributed graph to represent the human body. Specifically, we decompose the images into many overlapping patches, and represent each image with an undirected attributed graph \small $G=(V, E, A)$, where each vertex $v_{i}$ in the vertex set \small $V=\{v_i\}_{i=1}^{n}$ denotes an image patch, each edge encodes the contextual information of the connected vertex pair, and the vertex attributes \small $A = \{A^{P},A^{F}\}$ represent spatial and visual features of local patches.

During training, given a pair of positive images $I_1$ and $I_2$ with identity labels $l_1$ and $l_2$, where $l_1=l_2$ (i.e., $I_1$ and $I_2$ belong to the same person), they can be represented with attributed graphs $G_1=(V_1, E_1, A_1)$ and $G_2=(V_2, E_2, A_2)$, respectively. The patch-wise correspondence learning aims to establish the vertex correspondences $X \in \{0,1\}^{n_1\times n_2}$ between $V_1$ with $n_1$ vertices and $V_2$ with $n_2$ vertices, such that the intra-person matching score (i.e., $l_1 = l_2$) is maximized on the training set.

In Re-ID, $X_{i_1i_2}=1$ means the $i_1$-${th}$ patch in $I_1$ is matched with the $i_2$-${th}$ patch in $I_2$. We adopt the formulation in Eq.~\ref{eq:concise_formulation} to model our patch-wise correspondence learning problem:
\begin{equation}\small
\begin{aligned}
& \arg \mathrm{\max\limits_\mathbf{x}} {\kern 1pt}{\kern 1pt}{\kern 1pt}{\kern 1pt}  {\kern 1pt}{\kern 1pt} \mathbf{x}^{\mathrm{T}}K\mathbf{x}, \\
s.t.& \begin{cases}
 {\kern 1pt}{\kern 1pt}{\kern 1pt} &X_{i_1i_2}\in\{0,1\} ,\forall i_1\in\{1,\cdots,n_1\} ,\forall i_2\in\{1,\cdots,n_2\} \\
&{\sum}_{i_1} X_{i_1i_2} \leq 1, \forall i_2\in\{1,\cdots,n_2\},\\
&{\sum}_{i_2} X_{i_1i_2} \leq 1, \forall i_1\in\{1,\cdots,n_1\},
\end{cases}
\end{aligned}
\label{eq:graph_matching_obj}
\end{equation}
where one-to-one matching constraints are imposed on the assignment matrix $X$. We follow~\cite{RRWM/ChoLL10} to optimize Eq.~\ref{eq:graph_matching_obj}.

\subsubsection{\bf{Affinity matrix design}}

Due to the large variations in human body configurations caused by heavy pose and view changes, it is not suitable to directly apply traditional spatial layout based affinity matrix for Re-ID. In addition, taking into consideration the importance of visual appearance in Re-ID, we combine both visual feature and spatial layout of human appearance to develop the affinity matrix.

In specific, the diagonal components $K_{i_1i_2,i_1i_2}$ of the affinity matrix $K$ (which capture the node compatibility between vertex $v_{i_1} \in V_1$ and vertex $v_{i_2} \in V_2$) are calculated as follows:

\begin{equation}\small
K_{i_1i_2,i_1i_2} = S_{i_1i_2}^P \cdot S_{i_1i_2}^F,
\end{equation}
where $S_{i_1i_2}^P$ and $S_{i_1i_2}^F$ refer to the {\it spatial proximity} and {\it visual similarity} between $v_{i_1}$ and $v_{i_2}$ respectively. The $S_{i_1i_2}^P$ and $S_{i_1i_2}^F$ can be mathematically computed as:
\begin{equation}\small
\begin{aligned}
S_{i_1i_2}^P = \mathrm{exp}(-\| A_{i_1}^{P} - A_{i_2}^{P} \|_{2}), \\
S_{i_1i_2}^F = \mathrm{exp}(-\| A_{i_1}^{F} - A_{i_2}^{F} \|_{2}),
\end{aligned}
\end{equation}
where $A_{i_1}^{P}$ and $A_{i_2}^{P}$ denote spatial positions of $v_{i_1}$ and $v_{i_2}$, and $A_{i_1}^{F}$ and $A_{i_2}^{F}$ represent their visual features.

Likewise, for non-diagonal element $K_{i_1i_2,j_1j_2}$ in $K$, which encodes the compatibility between two matched vertex pairs $(v_{i_1} \in V_1,v_{i_2}\in V_2)$ and $(v_{j_1} \in V_1,v_{j_2} \in V_2)$, it can be obtained as the following:
\begin{equation}\small
K_{i_1i_2,j_1j_2} = S_{i_1j_1,i_2j_2}^P \cdot S_{i_1j_1,i_2j_2}^F,
\end{equation}
where $S_{i_1j_1,i_2j_2}^P$ and $S_{i_1j_1,i_2j_2}^F$ represent spatial and visual compatibilities between matched patch pairs $(v_{i_1} ,v_{i_2})$ and $(v_{j_1} ,v_{j_2} )$, and they are calculated by
\begin{equation}\small
\begin{aligned}
S_{i_1j_1,i_2j_2}^P = \mathrm{exp}(-\| (A_{i_1}^{P} - A_{j_1}^{P}) - (A_{i_2}^{P} - A_{j_2}^{P}) \|_{2}), \\
S_{i_1j_1,i_2j_2}^F = \mathrm{exp}(-\| (A_{i_1}^{F} - A_{j_1}^{F}) - (A_{i_2}^{F} - A_{j_2}^{F}) \|_{2}).
\end{aligned}
\end{equation}
In this way, the calculated affinity matrix $K$ implicitly embeds the spatial and visual contextual information into the graph matching procedure, such that during the optimization, correspondences with larger node similarities and more compatible edges are selected. Therefore, we can obtain a spatially and visually compatible patch-wise matching result for Re-ID.
\begin{figure}[!t]
  \centering
  \includegraphics[width=\linewidth]{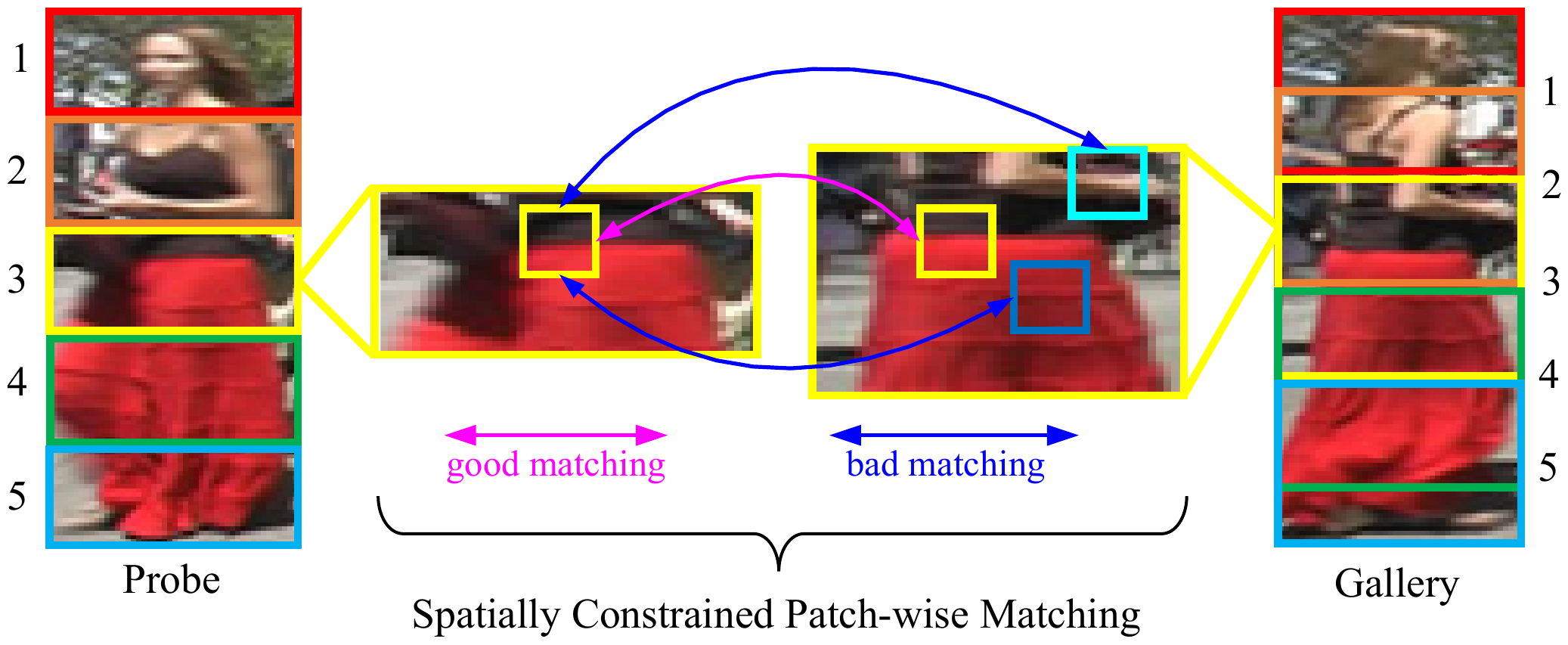}
 %    \captionsetup{font={footnotesize}}
  \caption{Illustration of the spatially constrained patch-wise matching. Following the commonly utilized stripe decomposition of the human body~\cite{sim_spatial_constraints}, we first divide the probe image into a few stripes, then the search space for each probe stripe is spatially constrained to its local neighborhood (e.g., patches within the yellow stripe of the probe image is constrained to be matched with patches in the corresponding yellow stripe of the gallery image). To ensure that the counterpart patch exists in the search space, the corresponding gallery stripe is set to be larger than the probe stripe. Best viewed in color.}
%  \vspace{-3ex}
  \label{fig::spatial_constraint}
\end{figure}
%\begin{figure}[!t]
 % \centering
 % \includegraphics[width=0.95\linewidth]{invisible.pdf}
 %  %  \captionsetup{font={footnotesize}}
 % \caption{Sample images to demonstrate the fact that local patches visible in one view may %not appear in the other.}
 % %\vspace{-2ex}
 % \label{fig::invisible}
%\end{figure}

\subsubsection{\bf{Spatially constrained matching}}
In existing part-based Re-ID methods~\cite{iccv15_correspondence,salicency}, an image is typically decomposed into hundreds of patches to capture detailed local visual information, leading to intractability in solving Eq. (\ref{eq:graph_matching_obj}). To reduce the search space and inhibit potential matching ambiguity, similar to the commonly utilized spatial constraints~\cite{lomo,sim_spatial_constraints}, we introduce the structure constrained matching. More specifically, a probe image is divided into a few horizontal stripes (non-overlapping), and for each probe stripe, its search space for patch-wise matching is constrained to a corresponding stripe in the gallery image (As shown in Fig.~\ref{fig::spatial_constraint}). Then patch-wise matchings are established between the probe stripe and the corresponding gallery stripe by optimizing Eq. (\ref{eq:graph_matching_obj}). Fig.~\ref{fig::spatial_constraint} illustrates the process of spatially constrained patch-wise matching.

%By optimizing Eq.(\ref{eq:graph_matching_obj_final}), we can obtain a set of graph correspondence templates for the positive image pairs in the training set.

\subsection{Reference selection via pose-pair configuration comparison}
\label{ref_selection}
We argue that the learned patch-wise correspondence patterns can be favorably transferred to image pairs with similar pose-pair configurations in the testing set, and these transferred correspondences can be directly utilized to compute the distance between probe and gallery images in the test set (as demonstrated in Fig.~\ref{fig:observation}). To this end, we need to find out the best references for each test pair from the training set.

In our preliminary study~\cite{our_gct/Zhou18}, we adopt body orientations to roughly capture body configuration. However the representative ability of body orientation is limited. In this paper, we propose a novel pose context descriptor to model the relative spatial distributions of the body joints to fully capture the spatial layout of human body. Please note here the locations of the body joints can be easily obtained with off-the-shelf pose estimation algorithms (e.g.,~\cite{real_time_pose_estimation/CaoSWS17}). For self-completeness, we present both the orientation based and the pose context descriptor based body configuration comparison strategies and compare their recognition performance in the experimental part.
\subsubsection{\bf{Comparing body configurations using body orientation}}

\begin{figure}[!t]
  \centering
  \includegraphics[width=\linewidth]{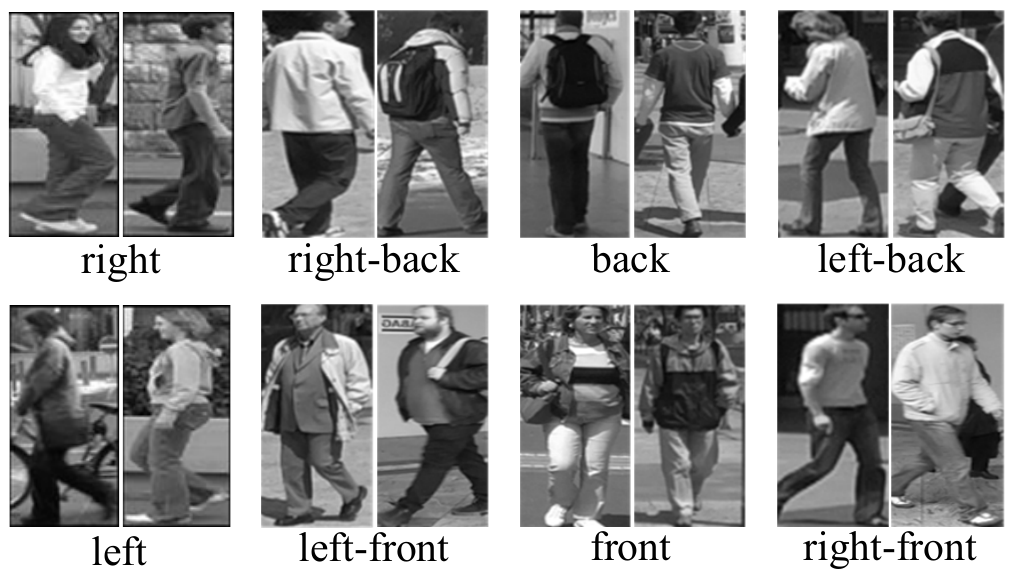}
%  \captionsetup{font={footnotesize}}
  \caption{Sample images in eight classes in the TUD dataset~\cite{TUD_DATASET}. Note that the TUD dataset is only used for training the body orientation classification model, and is different from the benchmark datasets in our experiments.}
%   \vspace{-3ex}
  \label{fig:multi_view_img}
\end{figure}

%\footnote{Due to limited space, please refer to~\cite{random_forest} for the details to build the random forest.}
We propose to utilize a simple yet effective random forest method~\cite{random_forest} to compare different body orientations. Specifically, images are classified into eight different clusters including `left', `right', `front', `back', `left-front', `right-front', `left-back' and `right-back', according to their body orientations, as shown in Fig.~\ref{fig:multi_view_img}. In order to train the random forest model, each image is represented with multi-level HoG features (i.e., cell sizes are set to $8 \times 8, 16\times 16, 32\times 32$ respectively, with a block size of $2 \times 2$ cells and a block stride of one cell for each direction), and then fed into each decision tree to build the random forest~\cite{random_forest}. Once the random forest $\mathcal{M} = \{M_i\}_{i=1}^{|\mathcal{M}|}$ is built, where $|\mathcal{M}|$ denotes the number of trees in $\mathcal{M}$, the body configuration similarity $\mathrm{O}$ between two images $I_i$ and $I_j$ can be calculated as:
\begin{equation}\small
\mathrm{O}(I_i,I_j) = \frac {1}{|\mathcal{M}|}\sum\limits_{m = 1}^{|\mathcal{M}|}{y_{ij}^m},
\label{eq:orientaiton_proximity}
\end{equation}
where $y_{ij}^m$ is an indicator, and $y_{ij}^m = 1$ if $I_i$ and $I_j$ arrive at the same terminal node in $M_m \in \mathcal{M}$, otherwise $y_{ij}^m = 0$.

\subsubsection{\bf{Comparing body configurations using pose context descriptor}}
%\IEEEpubidadjcol
To accurately model the human body configurations, we propose a novel pose context descriptor, which captures the relative spatial distribution across different joint pairs. In detail, fourteen body joints (i.e., head, neck, left/right shoulders, left/right elbows, left/right wrists, left/right coxaes, left/eight knees and left/right ankles) are estimated with the pre-trained model in~\cite{real_time_pose_estimation/CaoSWS17}, the spatial locations of which are denoted as $\{J_i, i\in 1,\cdots,14\}$. Each joint is then associated with a local polar coordinate system, centering at $J_i$. Then for the other joints with locations $\{J_j,j\in\{1,\cdots,14\},j\neq i\}$, we calculate a histogram for each $J_j$ by considering the magnitudes and angles of $J_j$ in the polar system centered at $J_i$. In this way, two pose context coding matrices $\Psi \in R^{14 \times 13},\Phi \in R^{14 \times 13}$ are formed, where the element $\Psi(i,j)$ in row $i$ column $j$ of $\Psi$ contains the \textbf{magnitude bin} that $J_j$ lies in the polar system centered at $J_i$. Likewise, $\Phi(i,j)$ is the \textbf{angle bin} that $J_j$ lies in the polar system centered at $J_i$. In this paper, the number of bins are set to 8 for both the magnitude and the angle. For a pair of images $I_1,I_2$, their pose configuration similarity is calculated as:
 \begin{equation}\small
\mathrm{O}(I_1,I_2) = S_{\Psi}(\Psi_1,\Psi_2)\cdot S_{\Phi}(\Phi_1,\Phi_2),
\label{eq:orientaiton_proximity1}
\end{equation}
where $S_{\Psi}(\Psi_1,\Psi_2)$ is the similarity score between the magnitude context matrices $\Psi_1$ and $\Psi_2$, and $S_{\Phi}(\Phi_1,\Phi_2) $ is the similarity score between the angle context matrices $\Phi_1$ and $\Phi_2$, which are calculated as follows:
\begin{equation}\small
S_{\Psi}(\Psi_1,\Psi_2) = \mathrm{mean}(\mathrm{exp}(-d(\Psi_1,\Psi_2))),
\end{equation}
where $d(\cdot)$ is the element-wise Euclidean distance. Considering that the angle bins are cyclic (e.g., $0$ degree equals to $360$ degree), we design a cyclic Euclidean distance to capture the proximity between angle bins,
\begin{equation}\small
S_{\Phi}(\Phi_1,\Phi_2) = \mathrm{mean}(\mathrm{exp}(-d_c(\Phi_1,\Phi_2))),
\end{equation}
 where $d_c(\cdot)$ is the cyclic Euclidean distance. For a pair of counterpart angle bins $(\Phi_1(i,j) \in \{1,\cdots,8\}, \Phi_2(i,j)\in \{1,\cdots,8\})$, denote the minimum bins from $\Phi_1(i,j)$ to $\Phi_2(i,j)$ in the angle circle as $\alpha_{ij}$, then $d_c(\Phi_1(i,j),\Phi_2(i,j))$ is calculated as $d_c(\Phi_1(i,j),\Phi_2(i,j)) =\alpha_{ij}^2 $.

\begin{figure}[!t]
  \centering
  \includegraphics[width=\linewidth]{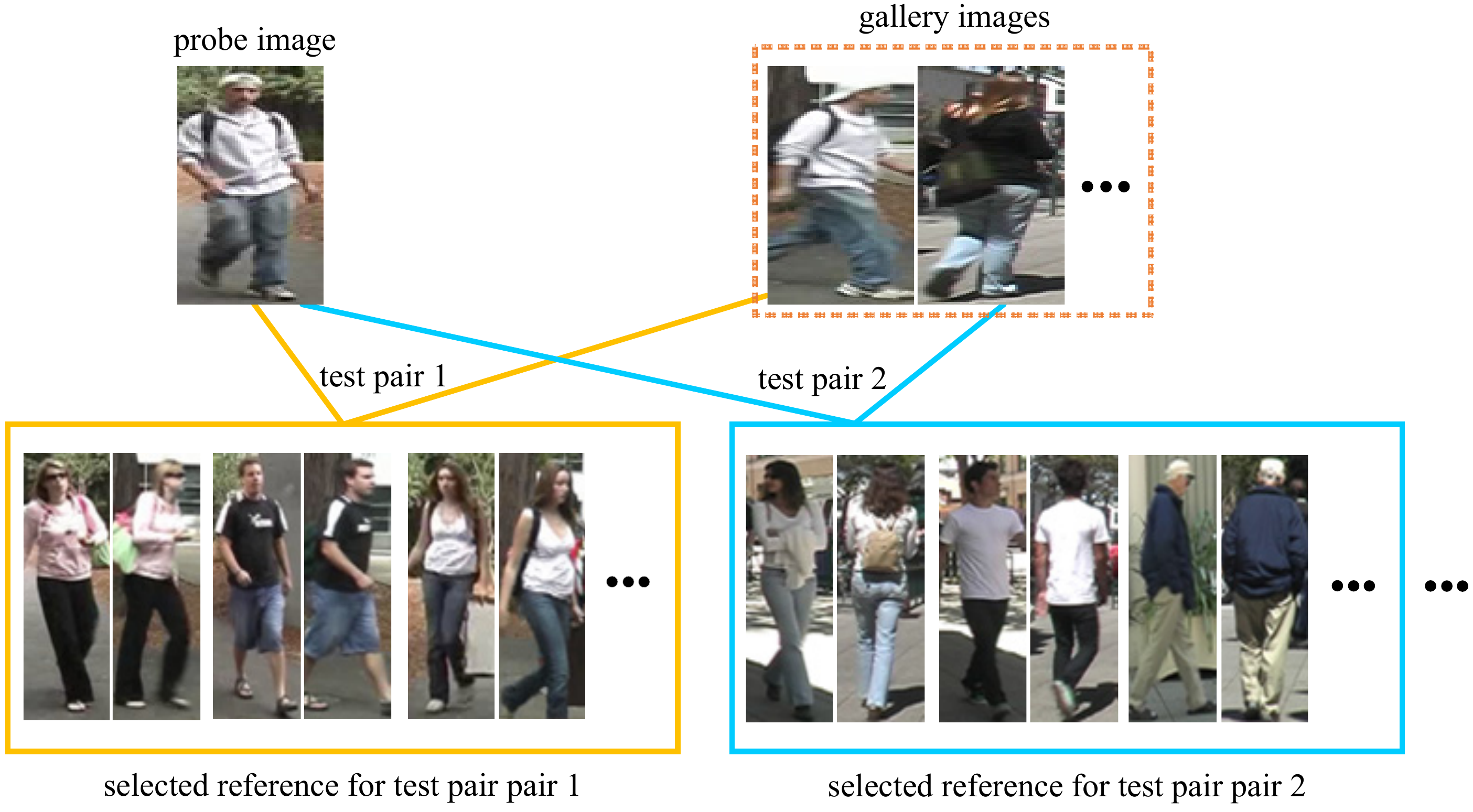}
%     \captionsetup{font={footnotesize}}
  \caption{Demonstration of reference selection results.}
 % \vspace{-3ex}
\label{fig:proximity_results}
\end{figure}

Given two image pairs $P=(I_p, I_g)$ and $P'=(I_p^{'}, I_g^{'})$, their pose-pair configuration similarity $S(P, P')$ is computed as
\begin{equation}\small
S_o(P, P') = \mathrm{O}(I_p,I_p^{'}) \cdot \mathrm{O}(I_g,I_g^{'}),
\label{eq:body_sim}
\end{equation}
where $\mathrm{O}(I_p,I_p^{'})$, $\mathrm{O}(I_g,I_g^{'})$ can be obtained using Eq.~(\ref{eq:orientaiton_proximity}) or Eq.~(\ref{eq:orientaiton_proximity1}).
With Eq.~(\ref{eq:body_sim}), we can calculate the body configuration similarities between an test image pair and all the positive training image pairs. Afterwards, $R$ training image pairs with highest similarities are selected as references for the test pair. And their learned patch-wise matchings are utilized for distance calculation as described later. Fig.~\ref{fig:proximity_results} shows some selected references of the sample test pairs.

\subsection{Distance calculation and aggregation with correspondence transfer}

Given that image pairs with similar pose-pair configurations tend to share similar patch-level correspondences, for each test pair of images, we propose to transfer the matching results of the selected references (the way to select the references is presented in Section~\ref{ref_selection}) to calculate the patch-wise feature distances of this test pair. The details of feature distance calculation using the selected references are presented in the following part.

Given a pair of test images $\bar{P}=(\bar{I}_p, \bar{I}_g)$, where $\bar{I}_p$ and $\bar{I}_g$ are the probe and gallery image respectively. Denote their corresponding graphs as $\bar{G}_p = (\bar{V}_p,\bar{E}_p,\bar{A}_p),\bar{G}_g = (\bar{V}_g,\bar{E}_g,\bar{A}_g)$, we can choose $R$ references for $\bar{P}$ as described in Section 3.2. Let $\mathcal{T}=\{ T_i \}_{i=1}^{R}$ represent the correspondence templates set composed by these $R$ references, where each template $T_{i}=\{c_{ij}\}_{j=1}^{n}$ contains $n$ patch-wise correspondences ($n$ is the number of graph node in the probe image), and each correspondence $c_{ij}=(w_{ij}^{p}, w_{ij}^{g})$ denotes the indices of the matched patches in the probe and gallery image (i.e., the $w_{ij}^{p}$-$th$ node in $\bar{G}_p$ is matched to the $w_{ij}^{g}$-$th$ node in $\bar{G}_g$).

For the test pair $\bar{P}$, we can compute the distance $\mathcal{D}$ between $\bar{I}_p$ and $\bar{I}_g$ as the following:
\begin{equation}\small
\mathcal{D}(\bar{I}_p, \bar{I}_g) = \frac{1}{R\times n} \sum\limits_{i=1}^{R}\sum\limits_{j=1}^{n}{\delta(\bar{A}_{p,w_{ij}^p},\bar{A}_{g,w_{ij}^g})},
\label{rank}
\end{equation}
where $\delta(\cdot,\cdot)$ denotes the distance metric (in this paper, we adopt the KISSME metric~\cite{KISSME}), $\bar{A}_{p,w_{ij}^p}$ and $\bar{A}_{g,w_{ij}^g}$ represent the visual attributes of the $w_{ij}^{p}$-$th$ patch in the probe image $\bar{I}_p$ and the $w_{ij}^{g}$-$th$ patch in the gallery image $\bar{I}_g$ respectively. In this paper, we use Local Maximal Occurrence features~\cite{lomo} as the visual attributes of each node (i.e., local patch).

With Eq.~(\ref{rank}), we can calculate the average patch-wise feature distance using all the correspondences (semantically matched patch pairs between the probe and gallery image) of the selected reference templates. For each probe image, the gallery image with the smallest distance is determined to be the re-identifying result.

\textbf{A more efficient solution for testing:}
As demonstrated in~\cite{our_gct/Zhou18}, for larger or more challenging datasets (e.g., CUHK01,VIPeR etc.), the number of selected reference templates $R$ is suggested to be large to accumulate enough correct patch-wise correspondences (e.g., $R$ is set to 20 for CUHK01 and VIPeR datasets). Therefore, according to Eq.~(\ref{rank}), for a dataset with $N$ persons (in the single-shot case, for $N$ persons, there are $N$ probe images and $N$ gallery images), the computational complexity involves $R\times n \times N^2$ patch-wise Mahalanobis distance calculations (here $n$ is the number of the graph node in the probe image).

In practical scenarios, it is desirable to have more efficient solutions. Therefore, we present a new evaluation protocol to significantly reduce the computational load with competitive or better performance. Since directly calculating the feature distances using all the reference patch-wise correspondences is time-consuming, we propose to aggregate the selected $R$ reference matching templates into $k, k\ll R$ more compact matching patterns via a voting scheme.

According to Eq.~(\ref{rank}), for each probe patch $w_{ij}^{p} \in \{1,\cdots,n\}$ in $\bar{I}_p$, the $i$-$th$ selected template matches it with patch $w_{ij}^{g}  \in \{1,\cdots,n\}$ in $\bar{I}_g$. Denote their spatial offset as:
 \begin{equation}\small
\Delta(w_{ij}^{p},w_{ij}^{g}) = \mathcal{L}(w_{ij}^{g})-\mathcal{L}(w_{ij}^{p}),
\label{offset_def}
\end{equation}
where $\mathcal{L}(w_{ij}^{p})$ denotes the center location for the $w_{ij}^{p}$-$th$ patch. Then for each probe patch $w_{ij}^{p}$, we can obtain a set of suggested matching patches $\Lambda(w_{ij}^{p}) = \{w_{ij}^{g},i \in 1,\cdots,R\}$. Assume these suggested matching patches vote for the hidden semantic matching $w_{ij}^{g*} \in \{1,\cdots,n\}$, then the location of $w_{ij}^{g*} $ can be simply derived as:
 \begin{equation}\small
 \mathcal{L}(w_{ij}^{g*})
 = \mathcal{L}(w_{ij}^{p})+ \frac{1}{R}\sum_{i=1}^R{\Delta(w_{ij}^{p},w_{ij}^{g})},w_{ij}^{g} \in \Lambda(w_{ij}^{p}),
\label{offset_mean}
\end{equation}
With the estimated target location, $k$ nearest patches in the target image are then sampled as the compact semantic matching patches of $w_{ij}^{p}$, which are calculated as:
 \begin{equation}\small
\{w_{ij}^{g*} \}_k =\mathcal{N}_k(\mathcal{L}(w_{ij}^{g*})),k\ll R,
\label{final_match}
\end{equation}
where $\mathcal{N}_k(\cdot)$ returns the indices of the $k$ nearest gallery patches with respect to the inside calculated target location. In this way, the computational load can be reduced to $k\times n \times N^2, k \ll R$, which is significant especially when $N$ is large.

The final feature distance between $\bar{I}_p, \bar{I}_g$ is then derived as,
\begin{equation}\footnotesize
\mathcal{D}(\bar{I}_p, \bar{I}_g) = \frac{1}{k\times n} \sum\limits_{i=1}^{k}\sum\limits_{j=1}^{n}{\delta(\bar{A}_{p,w_{ij}^p},\bar{A}_{g,w_{ij}^{g*}})},w_{ij}^{g*} \in \{w_{ij}^{g*} \}_k.
\label{final_dist}
\end{equation}

\section{Experimental Results}

In this section, we present the details of our experimental results. First we briefly clarify the experimental setup of the proposed graph correspondence transfer algorithm, then we elaborate on the ablative study of each component in the REGCT model to explore the importance of each part and give some insights on how to select the best parameter settings. Finally, we present the comparison results with some state-of-the-art algorithms on five challenging benchmarks, including the VIPeR~\cite{viper} dataset, the Road~\cite{iccv15_correspondence} dataset, the PRID450S~\cite{prid_450s} dataset, the 3DPES~\cite{three_dpes} dataset and the CUHK01~\cite{CUHK01} dataset.  %  Some sample images of each dataset are presented in Fig.~\ref{fig:sample_images}, from which we can see that various changes in illuminations, poses and views commonly occur in these datasets.

\begin{figure}[!t]
  \centering
  \includegraphics[width=\linewidth]{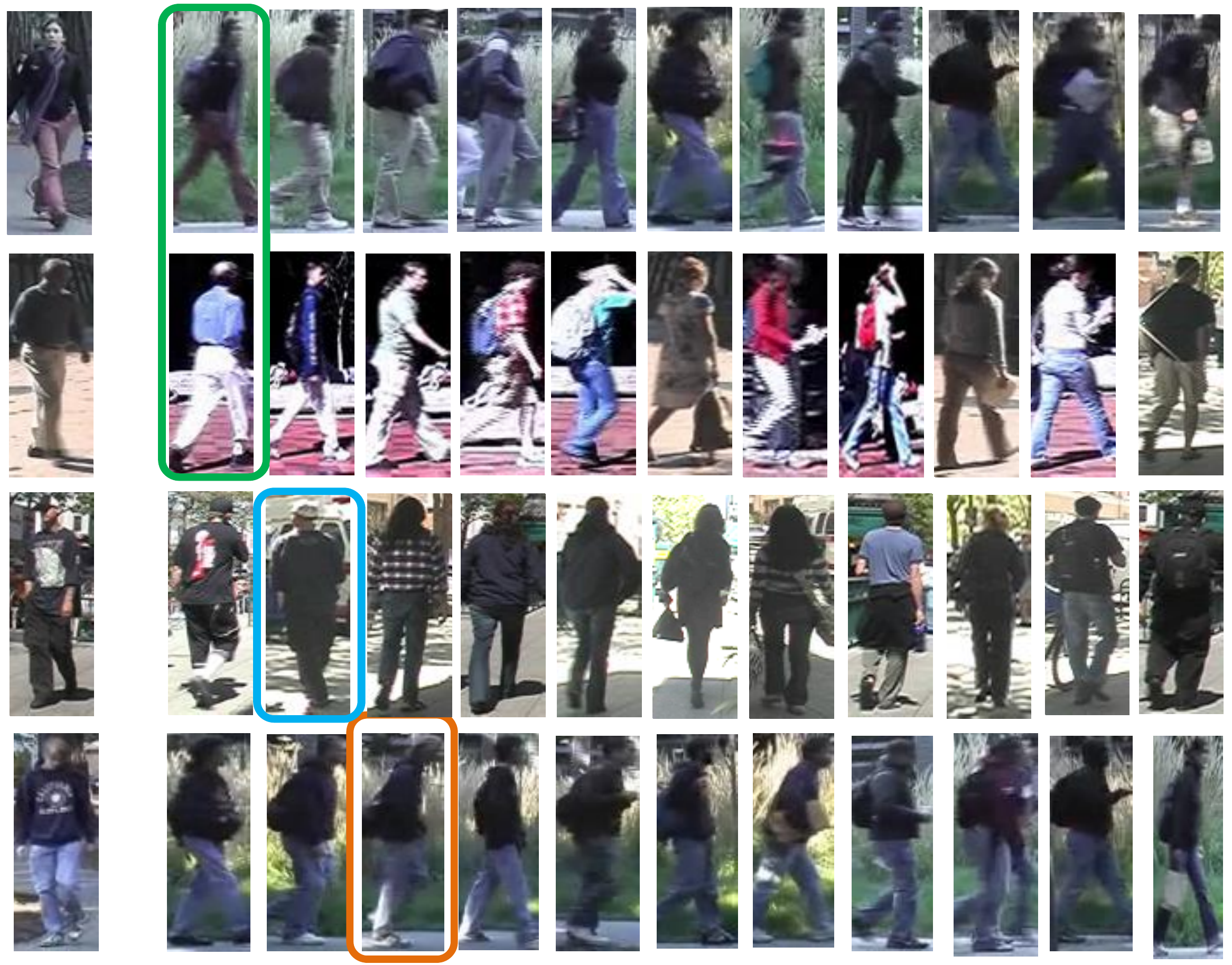}
%  \captionsetup{font={footnotesize}}
  \caption{Top ranked images selected by the proposed REGCT algorithm. Images in the first column are four randomly selected probe images, and the following are the top ranked gallery images of each probe by REGCT. Images marked by green/blue/orange bounding boxes in each row are the ground-truth matches of each probe.   }
%  \vspace{-3ex}
\label{fig:viper_results}
\end{figure}
\subsection{Experimental setup}
\subsubsection{\textbf{Datasets}}
We conduct experiments on three challenging single-shot datasets (VIPeR, Road and PRID450S), and two multi-shot datasets (3DPES and CUHK01). The characteristics of each dataset are detailed as follows:

\textbf{VIPeR dataset:} The VIPeR~\cite{viper} is a challenging person re-identification dataset consisting of 632 people with two images from two cameras for each person. It bears great variations in poses and illuminations, most of the image pairs contain viewpoint changes larger than 90 degrees.

\textbf{Road dataset:} The Road dataset~\cite{iccv15_correspondence}, consisting of 416 image pairs, is captured from a realistic crowd road scene, with serious interferences from occlusions and large pose variations, making it quite challenging.

\textbf{PRID450S dataset:} The PRID 450S~\cite{prid_450s} dataset contains 450 pairs of images from two camera views. The very similar background scene and many people wearing similar clothes make it very challenging for person re-identification.

\textbf{3DPES dataset:} The 3DPES dataset~\cite{three_dpes} contains 1011 images of 192 persons captured from 8 disjoint camera views, the images of which bear serious variations in view angles, illuminations, scales and background clutters. The number of images for a specific person ranges from 2 to 26, and the bounding boxes are generated from automatic pedestrian detection.

\textbf{CUHK01 dataset:} The CUHK01 dataset~\cite{CUHK01} is a medium-sized dataset for Re-id, captured from two disjoint camera views. It consists of 971 individuals, with each person having two images under each camera view. Different from VIPeR, images
in CUHK01 are of higher resolutions. On this dataset, we adopt the commonly utilized 485/486 setting for performance evaluation.

\subsubsection{\textbf{Parameter setup}} The proposed algorithm is implemented in Matlab on an Intel(R) Core(TM) i7-5820K CPU of 3.30GHz. The number of trees in the random forest model is 500.  The best configurations of $(R,k)$ as well as the impacts of different patch decompositions of human body are discussed in the following part. All the parameters are available in the source code at \url{http://www.dabi.temple.edu/~hbling/code/gct.htm}.

\begin{figure}[!t]
	\centering
	\includegraphics[width=\linewidth]{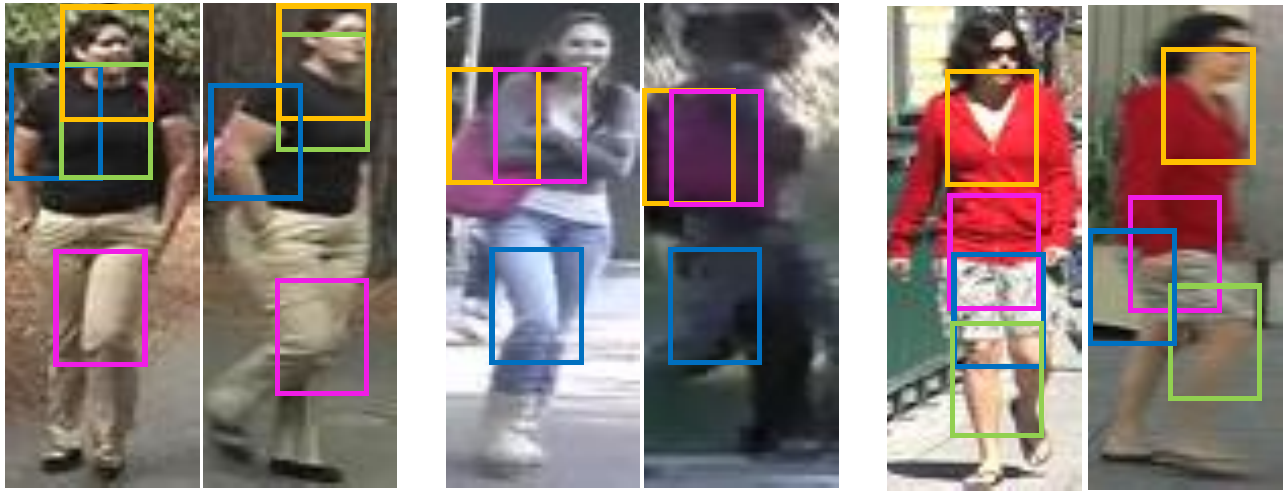}
	%  \captionsetup{font={footnotesize}}
	\caption{Some visualized patch-wise graph matching results. In each pair of images, the bounding boxes with the same color refer to an established correspondence by our algorithm.   }
	%  \vspace{-3ex}
	\label{fig:graph_matching_results}
\end{figure}

\noindent
\subsubsection{\textbf{Evaluation}} We adopt the commonly used half-training and half-testing setting~\cite{KISSME}, and randomly split the dataset into two equal subsets. The training/testing sets are further divided into the probe and gallery sets according to their view information. On all the datasets, both the training/testing set partition and probe/gallery set partition are performed 10 times and average performance is recorded. The performance is evaluated by cumulative matching characteristic (CMC) curve, which represents the expected probability of finding the correct match for a probe image in the top $r$ matches in the gallery list.

We record the top ranked gallery images of some sample probe images on the VIPeR dataset, which is presented in Fig.~\ref{fig:viper_results}. As shown in Fig.~\ref{fig:viper_results}, the proposed REGCT algorithm can successfully rank images visually similar to the probe image ahead of others, which is the key requirement of most existing surveillance systems. Please note the last row of Fig.~\ref{fig:viper_results}, the top ranked gallery images by REGCT are all with dark coats and blue jeans, and to some extent, the rank 1 image is visually more similar than the correct match marked by orange bounding box w.r.t. the probe image. Therefore, the proposed REGCT algorithm is able to handle the spatial misalignment problem and generate satisfying ranking results for practical applications.

%\begin{figure*}
%\centering
%\begin{subfigure}{.48\textwidth}
%  \centering
%  \includegraphics[width=1.05\linewidth]{patch_size}
%  \caption{Recognition performance with different patch size configurations.}
%  \label{fig:patch_wise_configa}
%\end{subfigure}%
%\begin{subfigure}{.48\textwidth}
%  \centering
%  \includegraphics[width=1.05\linewidth]{stride_config}
%  \caption{Recognition performance with different patch stride configurations.}
%  \label{fig:patch_wise_configb}
%\end{subfigure}
%\caption{Analysis on the impacts of different patch decompositions.}
%\label{fig:patch_wise_config}
% \vspace{-2ex}
%\end{figure*}

\subsection{Ablation study}
\subsubsection{\textbf{Visualized results of graph matching}}
To validate the effectiveness of graph matching in establishing local semantic correspondences, we present some patch-wise matching results in Fig.~\ref{fig:graph_matching_results}. As shown, by taking into consideration the spatial and visual context information during the graph matching procedure, the established patch-wise matchings indeed can preserve the semantic correspondences even if there are severe variations in poses and illuminations.
 %We present both the primary matching results and the matching results with refinement, from which we can see that the matched patch-wise correspondences with refinement are more consistent and visually more similar compared with the primary results.We also present the quantitative results to validate that the correspondence refinement step brings performance gain for person re-identification. The detailed comparison results are illustrated in Figure

\subsubsection{\textbf{Influence of different patch decomposition}}
In order to generate some insights into the optimal decomposition of human body into local patches, we study the influence of different patch-wise decompositions on the recognition performance. In specific, we record the recognition performance obtained using different patch sizes and different strides.

\begin{table*}[!tbp]\small
	%\captionsetup{font={footnotesize}}
	\caption{Recognition performance with different patch-wise configurations on VIPeR dataset. We study the influence of different configurations of both patch size and stride. Please note, in the patch size config, $h20\_w16$ indicates the patch height and width are set to 20 and 16 respectively. Likewise, in the patch stride config, $sh8\_sw6$ indicates that $stride\_h$ and $stride\_w$ are set to $8,6$ respectively. The best results are marked in red font.
		\label{tab:patch_wise_config}}
	\begin{center}
		\begin{tabular}{@{}C{2.7cm}@{}| @{}C{1.65cm}@{}|@{}C{1.65cm}@{}|@{}C{1.65cm}@{}|@{}C{1.65cm}@{}|@{}C{1.65cm}@{}|@{}C{1.65cm}@{}|@{}C{1.65cm}@{}|@{}C{1.65cm}@{}|@{}C{1.65cm}@{}}
			\hline
			patch\_size config  &h20\_w16 & h20\_w24 & h20\_w32 & h32\_w16 & h32\_w24 & h32\_w32 & h44\_w16 & h44\_w24 & h44\_w32 \\
			\hline
			\hline
			$rank=1$ & 42.2  &44.7 &49.1 & 45.7&48.9 & \textcolor{red}{\bf51.4}& 47.2& 49.0& 50.5\\
			$rank=5$& 73.2  & 75.9&77.7 &77.1 &\textcolor{red}{\bf80.8} &80.0 & 76.5&77.1 &79.1 \\
			$rank=10$ & 84.0  &86.0 &87.4 & 86.2& 87.1&\textcolor{red}{\bf88.1} & 86.2& 86.8&86.7 \\
			$rank=20$ & 93.7  &93.2 &93.9 &93.6 &94.0 & \textcolor{red}{\bf94.6}& 93.5& 93.3& 93.5\\
			
			\hline
			\hline
			stride config  &sh8\_sw6 & sh8\_sw8 & sh8\_sw12 & sh12\_sw6 & sh12\_sw8 & sh12\_sw12 & sh16\_sw6 & sh16\_sw8 & sh16\_sw12 \\
			\hline
			$rank=1 $& 49.2  &49.3 &47.1 & 49.5&\textcolor{red}{\bf49.7} &46.0 &48.4 &47.1 & 45.9\\
			$rank=5$& 77.0  &77.7 &75.8 &77.9 & \textcolor{red}{\bf78.1}& 75.4& 76.7&78.2 &74.7 \\
			$rank=10$ & 86.0  &86.4 &85.9 &86.6 &\textcolor{red}{\bf87.4} & 85.2& 86.0& 87.3& 85.7\\
			$rank=20$ &93.4   &92.9 & 92.8&93.0 &\textcolor{red}{\bf93.7} & 92.1& 92.6&\textcolor{red}{\bf93.7} &93.0 \\
			\hline
		\end{tabular}
	\end{center}
\end{table*}

\begin{figure*}[!t]
	\centering
	%\captionsetup{font={footnotesize}}
	\includegraphics[width=\linewidth]{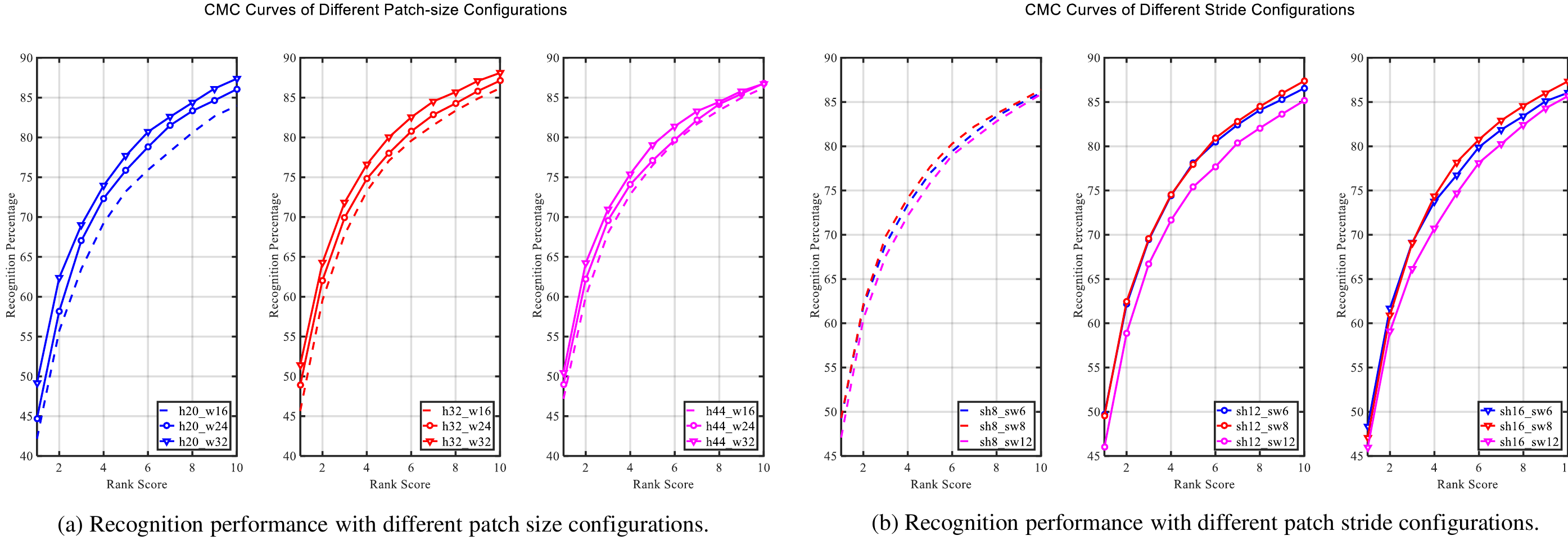}
	\caption{Analysis on the impacts of different patch decompositions.}
	\label{fig:patch_wise_config}
	%	\vspace{-2ex}
\end{figure*}

To evaluate the influence of patch size, we fix the stride to $stride\_h = 12, stride\_w = 8$ ($stride\_h$ is the stride along the height side, and $stride\_w$ is the stride along the width side). The patch height values and width values are then selected from $\{44,32,20\}$ and $\{32,24,16\}$ respectively, totally generating 9 combinations.
Likewise, we study the impact of different patch strides by fixing the patch size to $32 \times 24$, and then varying $stride\_h$ and $stride\_w$ within $\{16,12,8\}$ and $\{12,8,6\}$ respectively.  The detailed comparison results between different patch-wise decomposition settings (on the VIPeR dataset) are illustrated in Table~\ref{tab:patch_wise_config} and Fig.~\ref{fig:patch_wise_config}.

As shown in Table~\ref{tab:patch_wise_config}, the optimal patch size is $32\times32$, achieving the best performance at rank 1, 10, 20 and second best at $rank = 5$. As for the different stride configurations, stride $12\times 8$ performs the best at all the reported ranks among different stride configurations. Meanwhile, from Fig.~\ref{fig:patch_wise_config} (a) we can see that, with fixed patch height, the recognition performance increases when patch width gets larger. On the other hand, from Fig.~\ref{fig:patch_wise_config} (b), we observe that, best recognition performance is achieved at $stride\_w = 8$, when we fix $stride\_h$ to $8,12,16$ respectively.
\textbf{Based on the above analysis, we set the patch size to $32\times 32$ and strides to $stride\_h = 12, stride\_w = 8 $ throughout the following experiments without special clarification.}

\renewcommand\arraystretch{1}
\begin{table*}[!tbp]\footnotesize
	\centering
%\captionsetup{font={footnotesize}}
	\caption{Recognition performance of different $(R,k)$ configurations. In this table, each column contains the recognition rate at the same rank obtained with different $(R,k)$ configurations.  At each rank, the best and second best recognition rates are marked in red and blue respectively. Best viewed in color.}
	\label{fig:r_k} \begin{tabular}{p{0.1cm}p{0.2cm}|p{0.3cm}p{0.3cm}p{0.3cm}p{0.35cm}|p{0.3cm}p{0.3cm}p{0.3cm}p{0.35cm}|p{0.3cm}p{0.3cm}p{0.3cm}p{0.35cm}|p{0.3cm}p{0.3cm}p{0.3cm}p{0.35cm}|p{0.3cm}p{0.3cm}p{0.3cm}p{0.3cm}}

		\hline
		\hline \multicolumn{2}{c|}{\multirow{2}[0]{0.115\linewidth}{\diagbox[height=0.6cm]{{\scriptsize{{(R,k) }}}}{\scriptsize{{Datasets}}}}} & \multicolumn{4}{c|}{VIPeR}         & \multicolumn{4}{c|}{Road}         & \multicolumn{4}{c|}{PRID450S}         & \multicolumn{4}{c|}{3DPES}         & \multicolumn{4}{c}{CUHK01} \\
		\cline{3-22}
		\multicolumn{2}{c|}{} & r=1     & r=5    & r=10     & r=20     & r=1     & r=5     & r=10     & r=20    & r=1     & r=5    & r=10     & r=20     & r=1     &r=5     & r=10     & r=20     & r=1     & r=5     & r=10     & r=20 \\
		\hline
			\hline
		\multicolumn{1}{r}{\multirow{4}[0]{.05\linewidth}{R=10}} & k=1   & 49.9      &  78.3     & 87.2      &  94.2     &  86.4     &    95.3   &   97.5    &  99.0     &69.0       &  86.3     &    92.0   &   96.7    &  75.3     &  \textcolor{red}{\bf92.0 }    &    95.5   &   99.0    &  66.2     &   82.9    &89.8      & 93.7 \\ [-2 ex]
		& k=3   &51.6       & 80.1      &  88.3     & 94.5      & 88.5      & 96.0      &  97.7     &   \textcolor{blue}{\bf99.4}    &    \textcolor{blue}{\bf69.1}   &  \textcolor{blue}{\bf88.2}     & \textcolor{red}{\bf93.4}      &   \textcolor{red}{\bf97.0}    &  75.3     &   91.7   &   95.8    &  98.3     &   65.4    & \textcolor{blue}{\bf84.6 }     &  89.5     &94.2  \\[-2 ex]
		& k=5   & 51.2      &   80.0    &   88.3    &   94.3    &   88.2    & 95.5      &   98.2    &   99.3    &    67.8   &87.0       & 92.7      & 96.8      &  73.3     &  89.9     & \textcolor{blue}{\bf96.2   }   &   \textcolor{red}{\bf99.7 }   &  60.3     & 81.5      &   88.3   & 92.9 \\[-2 ex]
		& k=10   & 49.4      &   78.9    &    87.2   &   94.4    &  86.3     &    95.1   &   97.3    &98.5       & 60.4      &   83.6    &  90.1     &   95.6    &   72.9    &91.7       &     95.8  &  98.4     &   58.6    &   78.9    &  84.7     &  90.9\\
\hline
		\multicolumn{1}{r}{\multirow{4}[0]{.05\linewidth}{R=20}} & k=1   &  51.3     &  79.2     &   88.3    &   94.1    &   87.7    &  96.1     &   97.8    &    99.2   &   68.4    &    86.9   &  92.0    &    96.4   &   72.9   &  90.6    &     95.5  &   97.9   &  \textcolor{blue}{\bf66.7 }    &   83.8    &   \textcolor{blue}{\bf90.7}    &  93.9\\[-2 ex]
		& k=3   & 51.8      &  80.0     &    87.9   &  \textcolor{red}{\bf94.7}     &  \textcolor{blue}{\bf88.8 }    & 96.3      &  \textcolor{blue}{\bf98.5}     & \textcolor{red}{\bf99.6 }     & 69.5      &    87.9   & 92.9      & \textcolor{blue}{\bf96.9  }    & 76.0      & 91.7     &   95.1    &   98.3    &   64.3    &  83.5     &  89.3     & \textcolor{blue}{\bf94.5 }\\[-2 ex]
		& k=5   & 51.2      & 79.8      &   88.0    &   \textcolor{red}{\bf94.7}    &  88.6     & 96.1      &    98.1   &   99.3    &  67.4     &    87.5   &   92.8    & 96.5      & 75.7      &91.3     &   \textcolor{red}{\bf96.5 }   & 99.0      &  64.0     &  82.0     &  87.7     &92.7  \\[-2 ex]
		& k=10   & 49.1      &     80.0  &  87.2     & 94.2      &  86.8     & 95.3      &    97.3   &  98.8     &   63.8    &  85.5     &   91.2    &  96.2     &  71.7     &  88.5     &   94.3    &   97.7    &  58.5     & 78.9      &  84.8     & 91.0 \\
\hline
		\multicolumn{1}{r}{\multirow{4}[0]{.05\linewidth}{R=50}} & k=1   &  51.1     &  79.3     &   88.0    &  94.3     & 87.1      &  95.0     &  97.3     &   98.8    & 68.9      &    87.9   &    92.5   &   96.2    &\textcolor{red}{\bf77.1 }      &     \textcolor{red}{\bf92.0 }  &     \textcolor{red}{\bf96.5}  &   \textcolor{blue}{\bf99.3}    &   \textcolor{red}{\bf67.4}    &   \textcolor{red}{\bf86.2 }   &  \textcolor{red}{\bf90.9 }    & \textcolor{red}{\bf94.6} \\ [-2 ex]
		& k=3   & \textcolor{blue}{\bf52.5}      &   \textcolor{blue}{\bf80.3 }   &    \textcolor{blue}{\bf88.5}   &  94.5     &   87.0    & 95.6      &   98.1    &   \textcolor{blue}{\bf99.4}    &  67.2     &   86.6    & 91.8      &  96.0     &  \textcolor{blue}{\bf76.7 }    &  91.3     &   95.5    & 98.3      &  63.6     &   82.1    &     88.5  & 93.0 \\[-2 ex]
		& k=5   & 48.0      &    79.4   &    87.4   &   94.1    & 88.5      &95.3       &   97.6    &  \textcolor{blue}{\bf99.4}     & 67.7      &  86.3     &    92.5   &   95.9    &   68.4    &   89.9    &  95.1     &    97.9   &   61.7    &  81.6     & 88.5      &92.9  \\
		& k=10   &   49.2    &   78.4    &      87.3 &    93.9   & 87.8      &  95.4     & 97.8    &    99.3  &  60.6     &  84.3     & 91.1      &    96.1   &  75.0     &  88.0     &   94.3    &  97.9     & 58.6      &   79.0    &   85.2    &  91.0\\
\hline
		\multicolumn{1}{r}{\multirow{4}[0]{.05\linewidth}{R=100}} & k=1   &   51.2   &  80.0     &    88.0   &  94.4     &    86.7   &  95.0     &  97.2     & 98.5      &   69.1    &   86.7    &     92.2  &   96.4    &  73.3     &   90.6    &   95.5    &   97.6    &   66.5    &  84.5     &  90.0     & 94.2 \\
		& k=3   &  \textcolor{red}{\bf52.9}     &  \textcolor{red}{\bf80.5 }    &     \textcolor{red}{\bf88.7}  &  \textcolor{blue}{\bf94.6}     &   88.5   & \textcolor{red}{\bf96.5 }     &  \textcolor{red}{\bf98.6}     &  \textcolor{blue}{\bf99.4  }   &  \textcolor{red}{\bf69.7}     &    \textcolor{red}{\bf88.3}   &   \textcolor{blue}{\bf93.3 }   &   96.4    &  75.3     &  91.3     &    95.5   &    98.3   &  65.2     &   83.8    &  89.9     & 94.2 \\[-2 ex]
		& k=5   &  51.6     &    80.1   &   \textcolor{blue}{\bf88.5}    &   94.6    & \textcolor{red}{\bf89.0}     &95.9       &    98.1   &    99.2   &   66.6    &  86.6     &  92.4     &   96.6    &  71.9     &  89.9     &    93.8   &   97.6    &   61.6    &    81.9   &  88.1     & 93.4 \\
		& k=10   &   49.5    &   79.6    &  87.6     &  94.0     & 86.4      & 95.0      &   97.0    &  98.8     &  62.3     &   85.4    &  91.6     &  96.0     &   67.7    & 89.6      &   93.2    &   96.4    &  58.6     &   78.9    &  85.2     &  90.9\\
%\hline\hline
%\multicolumn{2}{r|}{ baseline GCT~\cite{our_gct/Zhou18} }  &  49.4 & 77.6 & 87.2 & 94.0 &  \textcolor{blue}{\bf 88.8} & \textcolor{red}{\bf 96.7} &  98.4 & \textcolor{red}{\bf 99.6}     &  58.4 &77.6 &  84.3 &  89.8    &     69.8 & \textcolor{red}{\bf 92.4} & 95.5 &  97.2  &   61.9 & 81.9 & 87.6 & 92.8\\
\hline
\end{tabular}
%\vspace{-3ex}
\end{table*}

\begin{figure*}[!t]
	\centering
	\includegraphics[width=\linewidth]{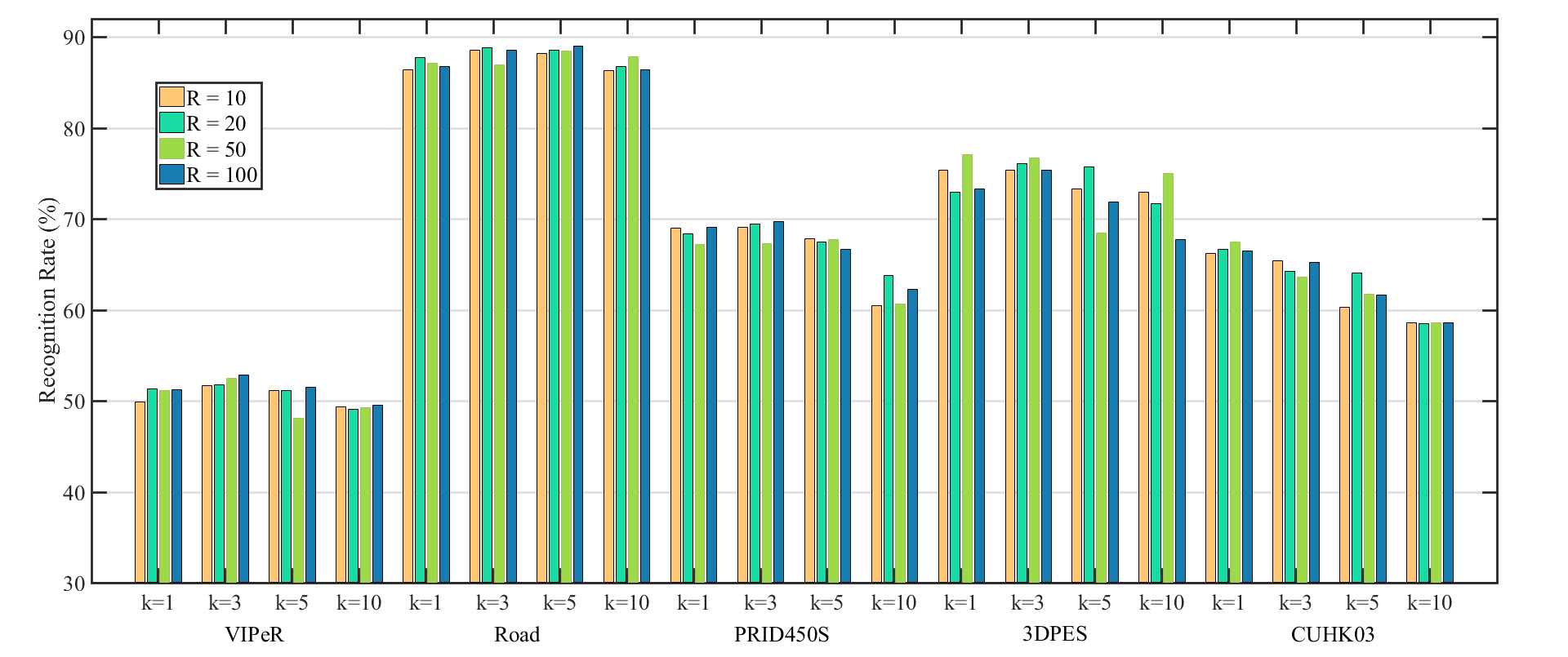}   % height=6cm
	%\captionsetup{font={footnotesize}}
	\caption{Influence of different $(R,k)$ settings on the rank-1 recognition performance on five benchmarks. Each group of bars illustrates the rank-1 performance at a fixed $k$ with varying $R$, and each dataset contains four successive groups of bars.    }
	% \vspace{-1ex}
	\label{fig:bar_rk}
\end{figure*}
\subsubsection{Analysis on different configurations of $(R,k)$}
The number of selected references ($R$) for calculating the distances between test pairs has an impact on the re-identification performance. With a small $R$, bad references may have a large impact on the patch-wise distance calculation, deteriorating the recognition performance. By contrast, if the value of $R$ is large, the correspondences transferred from less similar references may introduce inaccurate correspondences, which also degrades the performance. In the original evaluation setting~\cite{our_gct/Zhou18}, the optimal $R$ for the VIPeR, Road, PRID450S, 3DPES and CUHK01 datasets are 20, 5, 10, 20 and 20, respectively. In this paper, we propose to aggregate the arbitrary optimal number ($R$) of noisy reference templates into $k$ refined compact templates to improve the robustness of correspondence transfer as well as the testing efficiency. We conduct experiments to find out the best $(R,k)$ combination for all the benchmarks. More specifically, we set $R$ to $R \in\{10,20,50,100\}$, and vary $k$ within $\{1,3,5,10\}$, generating in total 16 configurations of $(R,k)$.

\renewcommand{\arraystretch}{1.2}
\begin{table*}[!t]

  \centering
 %  \captionsetup{font={footnotesize}}
 % \fontsize{6.5}{8}\selectfont
  \caption{Comparison results under different experimental settings to demonstrate the effectiveness of templates ensemble.}
  \label{tab:superiority_of_ensemble}
    \begin{tabular}{@{}C{2.7cm}@{}| @{}C{0.95cm}@{}|@{}C{0.95cm}@{}|@{}C{0.95cm}@{}|@{}C{0.95cm}@{}|@{}C{0.95cm}@{}|@{}C{0.95cm}@{}|@{}C{0.95cm}@{}|@{}C{0.95cm}@{}|@{}C{0.95cm}@{}|@{}C{0.95cm}@{}|@{}C{0.95cm}@{}|@{}C{0.95cm}@{}|@{}C{0.95cm}@{}|@{}C{0.95cm}@{}|@{}C{0.95cm}@{}}

    \hline\hline
    \multirow{2}{*}{\diagbox[height=0.7cm]{Settings}{Datasets}}&
    \multicolumn{3}{c|}{VIPeR}&\multicolumn{3}{c|}{ Road}&\multicolumn{3}{c|}{ PRID450S}&\multicolumn{3}{c|}{ 3DPES}&\multicolumn{3}{c}{ CUHK03}\cr\cline{2-16}
    &r=1&r=10&r=20&r=1&r=10&r=20&r=1&r=10&r=20&r=1&r=10&r=20&r=1&r=10&r=20\cr
    \hline
    \hline
     GCT$_{best}$  &  49.4  & 87.2 & 94.0 &   \textbf{88.8}  &  98.4 &\textbf{ 99.6}    &  60.0&89.1&94.6     &72.5&95.0&97.7   &62.2&88.9&93.5 \cr\hline
      REGCT$_{best}$&  \textbf{52.9}    &    \textbf{88.7}  &  \textbf{94.6}     &   88.5   &  \textbf{98.6}     &  99.4     &  \textbf{69.7}    &   \textbf{93.3 } &\textbf{96.4}&\textbf{77.1 }    &     \textbf{96.5}  &   \textbf{99.3}    &   \textbf{67.4}  &  \textbf{90.9 }    & \textbf{94.6} \cr\hline\hline
    GCT$_{one}$&46.0&85.1&93.1&83.1&96.3&98.8&58.4 &  84.3 &  89.8&     69.8  & 95.5 &  97.2&   61.9  & 87.6 & 92.8\cr\hline
   REGCT$_{one}$&   \textbf{51.2} &    \textbf{88.0}   &  \textbf{94.4 }  &  \textbf{ 87.7 }     &   \textbf{97.8}    &    \textbf{99.2} &  \textbf{ 69.1 }   &    \textbf{ 92.2 } & \textbf{  96.4 }  &\textbf{77.1 }    &     \textbf{96.5}  &   \textbf{99.3}    &   \textbf{67.4}    & \textbf{90.9 }    & \textbf{94.6}\cr\hline
    \end{tabular}
\end{table*}
 As shown in Table~\ref{fig:r_k}, generally, %for the aggregated compact matching templates setting,
 the best recognition performance are achieved at small $k$ with arbitrary optimal $R$. More specifically, $(R=100,k=3)$ outperforms other configurations on the VIPeR, Road and PRID450S datasets. While on the 3DPES and CUHK01 datasets, $(R=50,k=1)$ generates the best results. Another highly expected phenomenon is that large $k$ (e.g., set $k$ to 10) generally deteriorates the recognition performance regardless of the different settings of $R$. These two facts validate that the proposed template ensemble approach turns the multiple noisy reference templates into a more robust and accurate matching pattern. And the reason for the performance drop when using large $k$ can be inferred from Eq.~\ref{final_match}. In Eq.~(\ref{final_match}), the final patch-wise matching is selected according to the spatial proximity to the voted target location calculated by Eq.~(\ref{offset_mean}). When $k$ is large, more patches that deviate from the estimated target location are included in the distance calculation, thus leading to deterioration in the recognition performance.

As rank-1 performance is a very critical evaluation criterion for person re-identification, we specially study the influence of different $(R,k)$ settings on the rank-1 recognition rates. As shown in Fig.~\ref{fig:bar_rk}, generally, $k=3$ works best for the VIPeR, Road and PRID450S datasets, $k=1$ is optimal on the 3DPES and CUHK01 datasets. Compared with $k$, different configurations of $R$ have relatively small influence on the recognition rates. Since the number of $R$ does not affect the efficiency in the proposed novel evaluation setting, REGCT is guaranteed to obtain favorable performance with better efficiency than the baseline GCT~\cite{our_gct/Zhou18}.
\textbf{In the following part, the best $(R,k)$ configurations are utilized for each dataset without special clarification.}

\subsubsection{Analysis on the effectiveness of template ensemble}

As discussed earlier, the proposed REGCT can reduce the computational load from $R\times n \times N^2$ to $k\times n \times N^2$ ($k\ll R$), making REGCT more efficient than GCT when $N$ is large. In this subsection, we demonstrate that template ensemble in the proposed REGCT also brings improvements to the recognition performance. In particular, we record the best recognition rates obtained with optimal parameter settings using both REGCT and the baseline GCT (we denote the experimental settings as REGCT$_{best}$ and GCT$_{best}$ respectively). Besides, to further demonstrate the superiority of the aggregated compact templates over the original noisy templates, we record the performance obtained by using one ensembled template and one original template respectively (denoted as REGCT$_{one}$ and GCT$_{one}$). Please note for fair comparison, experiments under all the four settings are carried out with $(h,w,stride\_h,stride\_w)$ set to $(32,32,12,8)$ respectively. The detailed comparison results are recorded in Table~\ref{tab:superiority_of_ensemble}.

\renewcommand{\arraystretch}{1.2}
\begin{table*}[!t]

  \centering
 % \fontsize{6.5}{8}\selectfont
%  \captionsetup{font={footnotesize}}
  \caption{Comparison results of utilizing different body configuration comparison strategies for Re-ID. REGCT$_{best}$ records the results obtained by using the body orientation based strategy, while  REGCT$_{best}(_{w/p})$ reports the results of utilizing the pose context based strategy. }
  \label{tab:superiority_of_pose_context}
    \begin{tabular}{@{}C{2.7cm}@{}| @{}C{0.95cm}@{}|@{}C{0.95cm}@{}|@{}C{0.95cm}@{}|@{}C{0.95cm}@{}|@{}C{0.95cm}@{}|@{}C{0.95cm}@{}|@{}C{0.95cm}@{}|@{}C{0.95cm}@{}|@{}C{0.95cm}@{}|@{}C{0.95cm}@{}|@{}C{0.95cm}@{}|@{}C{0.95cm}@{}|@{}C{0.95cm}@{}|@{}C{0.95cm}@{}|@{}C{0.95cm}@{}}

    \hline\hline
    \multirow{2}{*}{\diagbox[height=0.7cm]{Settings}{Datasets}}&
    \multicolumn{3}{c|}{VIPeR}&\multicolumn{3}{c|}{ Road}&\multicolumn{3}{c|}{ PRID450S}&\multicolumn{3}{c|}{ 3DPES}&\multicolumn{3}{c}{ CUHK03}\cr\cline{2-16}
    &r=1&r=10&r=20&r=1&r=10&r=20&r=1&r=10&r=20&r=1&r=10&r=20&r=1&r=10&r=20\cr
    \hline
    \hline
    REGCT$_{best}$&  52.9    &   88.7  & 94.6     &   88.5   &  98.6     &  99.4  &  69.7  &   93.3  &96.4&77.1     &     \textbf{96.5}  &   \textbf{99.3}    &  67.4  &  90.9     & 94.6 \cr\hline
    REGCT$_{best}(_{w/p})$&  \textbf{53.5}    &    \textbf{89.1}  &  \textbf{94.7}     &   \textbf{88.9}   &  \textbf{98.9}     &  \textbf{99.6 }   &  \textbf{70.9}    &   \textbf{93.5 } &\textbf{96.5}&\textbf{78.0 }    &     \textbf{96.5}  &  99.1    &   \textbf{68.0}  &  \textbf{91.2 }    & \textbf{94.9} \cr\hline

      \end{tabular}
%      \vspace{-2ex}
\end{table*}

As shown in Table~\ref{tab:superiority_of_ensemble}, the best performance of REGCT$_{best}$ obtained using the optimal $(R,k)$ settings on each dataset consistently outperforms the corresponding best results recorded in GCT$_{best}$~\cite{our_gct/Zhou18}, except slightly lower but competitive rank-1/rank-20 recognition rates on the Road dataset. As for the one template setting, compared to GCT$_{one}$, REGCT$_{one}$ significantly boosts the performance at all ranks across all the five benchmarks. This obvious performance gain validate that the ensembled compact templates are more robust and accurate than the original noisy matching templates, therefore benefiting the recognition performance by a large margin. Besides, in REGCT, the results of REGCT$_{one}$ are fairly close to REGCT$_{best}$, which means competitive recognition performance can be obtained with only one reference template (therefore significantly reducing the computational load).
Please note that till now, all the experiments are recorded by using the body orientation based reference templates selection. The effectiveness of the proposed pose context descriptor based reference selection is discussed in the following part.
\begin{figure*}[!t]
  \centering
  \includegraphics[width=\linewidth]{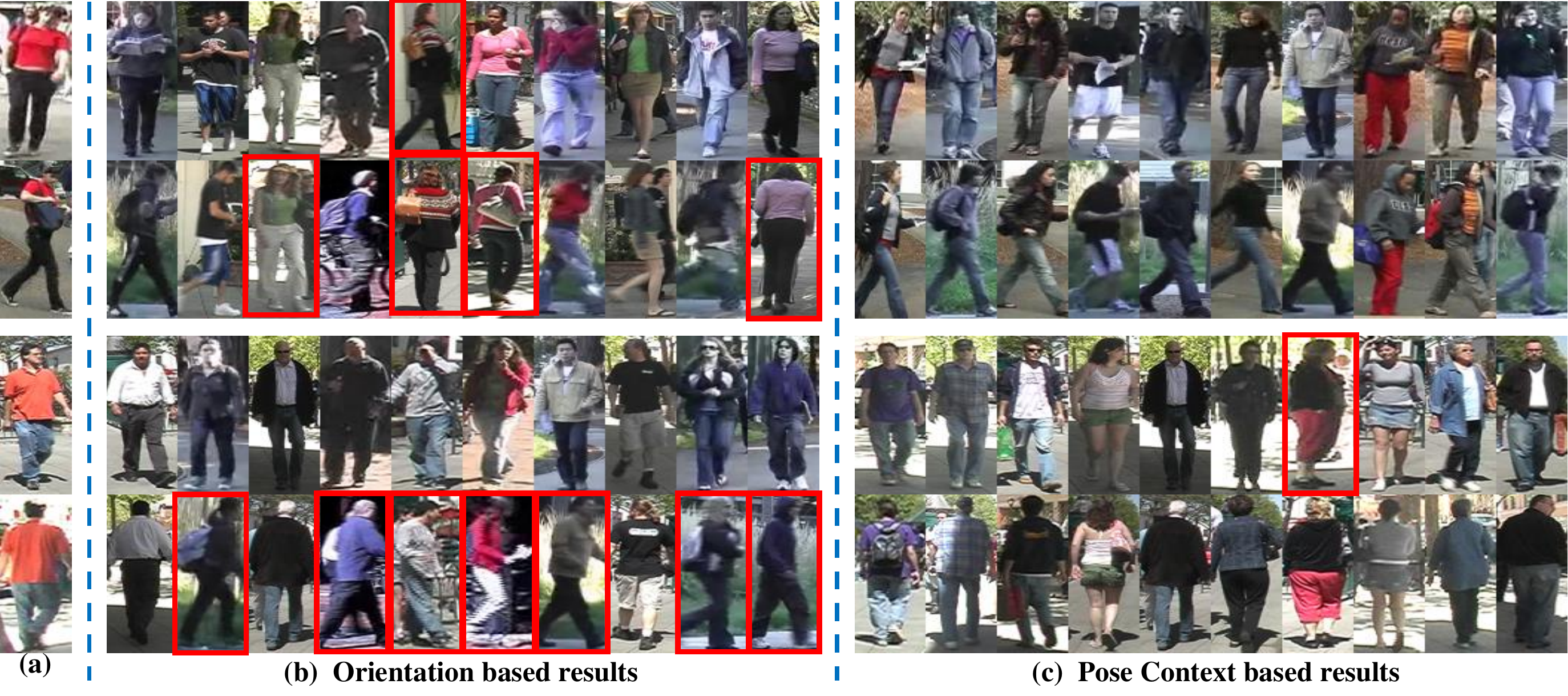}
 % \captionsetup{font={footnotesize}}
  \caption{The top-10 ranked reference image pairs for two sample test pairs. (a) Two sample test image pairs. (b) The top-10 selected reference positive training pairs generated by the orientation based method. (c) The top-10 ranked reference image pairs selected by the pose context descriptor based method. As shown, the context descriptor generates reference images with more consistent body configurations.  }
%  \vspace{-2ex}
\label{fig:pose_ref_results}

\end{figure*}

\subsubsection{\textbf{Evaluation on different body configuration comparison strategies}}
We also conduct experiments to demonstrate the superiority of the proposed novel pose context descriptor in modeling body configurations than the original orientation based method in~\cite{our_gct/Zhou18}. We present both the qualitative and the quantitative study as follows:

\textbf{Qualitative Study:} For each test pair, we record the top-10 ranked reference pairs according to their pose-pair configuration similarities calculated by both the orientation based and the pose context based methods. As shown in Fig.~\ref{fig:pose_ref_results}, the orientation based method generates more noisy reference images (marked by the red bounding boxes) than the pose context based results.

\textbf{Quantitative Study:} We also record the recognition performance based on the two different reference selection strategies. The detailed comparison results on challenging datasets are presented in Table~\ref{tab:superiority_of_pose_context}. As illustrated in Table~\ref{tab:superiority_of_pose_context}, the pose context descriptor based reference selection can indeed boost the Re-ID performance. But the relatively slight performance gain indicate that the proposed REGCT algorithm is robust to different templates selection strategies. This indicates that we can save the efforts of trying to build more accurate body configuration descriptors, but pay more attention to improving the matching accuracy of the training templates in the future research.

\begin{table}[!t]
% \captionsetup{font={footnotesize}}
\caption{Comparisons of top $r$ matching rate using CMC (\%) on VIPeR dataset. The best result is marked in red font, and the second best in blue.
\label{tab:VIPeRDataset}}
\begin{center}
\begin{tabular}{@{}C{3cm}@{}| @{}C{1.2cm}@{}@{}C{1.2cm}@{}@{}C{1.2cm}@{}@{}C{1.2cm}@{}}
\hline
 Methods  & r=1 & r=5 & r=10 & r=20 \\
 \hline
 \hline
SalMatch~\cite{salicency} &  $30.2$ & $52.3$ & $65.5$ & $79.2$\\
Semantic~\cite{Semantic15} &  $41.6$ & $71.9$ & $86.2$ & $\textcolor{blue}{\bf 95.1}$\\
LSSCDL~\cite{LSSCDL16} &  $42.7$ & $-$& $84.3$  & $91.2$\\
KISSME~\cite{KISSME} & $27.3$ & $55.3$ & $69.0$ & $82.7$\\
SVMML~\cite{SVMML} & $30.0$ & $64.7$ & $79.0$ & $91.3$\\
kLFDA~\cite{kLFDA14} &  $32.4$ & $65.9$ & $79.8$ & $90.8$\\
Polymap~\cite{ChenYHZW15} & $36.8$ &$70.4$ &$83.7$ & $91.7$ \\
LMF+LADF~\cite{mid_filter14} & $43.4$ & $73.0$ & $84.9$ & $93.7$ \\
LOMO+XQDA~\cite{lomo} &  $40.0$ & $68.1$ & $80.5$ & $91.1$\\
DCSL~\cite{DCSL} &  $44.6$ & $73.4$ & $82.6$ & $-$\\
TMA~\cite{TMA}&  $48.2$ & $-$ & $87.7$ & $\textcolor{red}{\bf 95.5}$\\
TCP~\cite{DEEP_MULTI_CHANNEL} &  $47.8$ & $74.7$ & $84.8$ & $91.1$\\
DGD~\cite{DGD} &  $35.4$ & $62.3$ & $69.3$ & $-$\\
Spindle-Net~\cite{Spindle_Net} &  $\textcolor{red}{\bf 53.8}$ & $74.1$ & $83.2$ & $92.1$\\
DML$^2$V~\cite{dml2v/SunWL17} & 50.4 & \textcolor{blue}{\bf80.5} & \textcolor{blue}{\bf88.7} & 95.0 \\
CSL~\cite{iccv15_correspondence} &  $34.8$ & $68.7$ & $82.3$ & $91.8$\\
\hline
\hline
REGCT$_{best}(_{w/p})$&  $\textcolor{blue}{\bf 53.5}$ & $\textcolor{red}{\bf 81.3}$ & $\textcolor{red}{\bf 89.1}$ & $94.7$\\\hline
\end{tabular}
\vspace{-3ex}
\end{center}
\end{table}
%\vspace{-1ex}
\subsection{\textbf{Comparison with state-of-the-arts}}

To demonstrate the effectiveness of the proposed approach, we compare the proposed REGCT with some state-of-the-art approaches on five challenging datasets. Please note results generated by REGCT$_{best}(_{w/p})$ are utilized in the following comparisons. And the detailed comparison results are presented as follows.

On the VIPeR dataset, we compare the REGCT with other sixteen algorithms, including SalMatch~\cite{salicency}, Semantic~\cite{Semantic15}, LSSCDL~\cite{LSSCDL16}, KISSME~\cite{KISSME}, SVMML~\cite{SVMML}, kLFDA~\cite{kLFDA14}, Polymap~\cite{ChenYHZW15}, LMF+LADF~\cite{mid_filter14}, LOMO+XQDA~\cite{lomo}, DCSL~\cite{DCSL}, TMA~\cite{TMA}, TCP~\cite{DEEP_MULTI_CHANNEL}, DGD~\cite{DGD}, Spindle Net~\cite{Spindle_Net}, DML$^2$V~\cite{dml2v/SunWL17} and CSL~\cite{iccv15_correspondence}. The comparison results are presented in Table~\ref{tab:VIPeRDataset}. As illustrated in Table~\ref{tab:VIPeRDataset}, the proposed REGCT algorithm achieves the best recognition rate at rank 5, 10, and competitive performances at rank 1, 20. Please note DCSL~\cite{DCSL}, TCP~\cite{DEEP_MULTI_CHANNEL}, DGD~\cite{DGD}, and Spindle Net~\cite{Spindle_Net} are deep feature based end-to-end framework, the favorable performance of REGCT against these algorithms demonstrates the effectiveness of our system for Re-ID. Besides, compared with CSL~\cite{iccv15_correspondence} and DCSL~\cite{DCSL}, which also aim to establish local semantic correspondences, our REGCT evidences notable performance gain. This validates the superiority of exploiting contextual information via graph matching to address the spatial misalignment problem.

\begin{table}[!htbp]
% \captionsetup{font={footnotesize}}
\caption{Comparison of top $r$ matching rate using CMC (\%) on Road dataset. The best result is marked in red font, and the second best in blue.
\label{Road_Dataset.}}
\begin{center}
\begin{tabular}{@{}C{3cm}@{}| @{}C{1.2cm}@{}@{}C{1.2cm}@{}@{}C{1.2cm}@{}@{}C{1.2cm}@{}}
\hline
 Methods  & r=1 & r=5 & r=10 & r=20 \\
 \hline\hline
eSDC-knn~\cite{salicency} &  $52.4$ & $74.5$ & $83.7$ & $89.9$\\
CSL~\cite{iccv15_correspondence} &  $\textcolor{blue}{\bf 61.5}$ & $\textcolor{blue}{\bf91.8}$ & $\textcolor{blue}{\bf 95.2}$ & $\textcolor{blue}{\bf98.6}$\\
\hline\hline
REGCT$_{best}(_{w/p})$ &  $\textcolor{red}{\bf 88.9}$ & $\textcolor{red}{\bf 96.8}$ & $\textcolor{red}{\bf 98.9}$ & $\textcolor{red}{\bf 99.6}$\\
\hline
\end{tabular}
\end{center}
%\vspace{-2ex}
\end{table}

The Road dataset is proposed in CSL~\cite{iccv15_correspondence}. For comprehensive comparison, we also report the result on this dataset, and compare it with eSDC-knn~\cite{salicency} and CSL~\cite{iccv15_correspondence}. As shown in Table~\ref{Road_Dataset.}, compared to CSL~\cite{iccv15_correspondence}, our algorithm obtains significant improvements of 27.4\%, 5.0\%, 3.7\%, 1.0\% at rank 1, 5, 10, 20 respectively. Owing to the sample-specific patch-wise matching adopted in our algorithm, significant performance gain is achieved compared with the camera-specific global matching structure adopted in CSL~\cite{iccv15_correspondence}.

On the PRID450S dataset, we compare with KISSME~\cite{KISSME}, SCNCDFinal~\cite{YangYYLYL14}, Semantic~\cite{Semantic15}, TMA~\cite{TMA}, NSFT~\cite{NSFT}, DML$^2$V~\cite{dml2v/SunWL17} and CSL~\cite{iccv15_correspondence}. As shown in Table~\ref{tab:prid_Dataset.}, our algorithm achieves the best recognition performance at all ranks. This obvious performance gain can be attributed to the accurate patch-wise correspondence via graph matching as well as the robust correspondence transfer via template ensemble.

\begin{table}[!hbtp]\small
 \captionsetup{font={footnotesize}}
\caption{Comparison of top $r$ matching rate using CMC (\%) on the PRID450S dataset. The best result is marked in red font, and the second best in blue.
\label{tab:prid_Dataset.}}
\begin{center}
\begin{tabular}{@{}C{3cm}@{}| @{}C{1.2cm}@{}@{}C{1.2cm}@{}@{}C{1.2cm}@{}@{}C{1.2cm}@{}}
\hline
Methods  & r=1 & r=5 & r=10 & r=20 \\
\hline\hline
KISSME~\cite{KISSME} &  $33$ & $-$ & $71$ & $79$\\
SCNCDFinal~\cite{YangYYLYL14}& $41.6$ & $68.9$ & $79.4$ & $87.8$\\
Semantic~\cite{Semantic15} &  $44.9$ & $71.7$ & $77.5$ & $86.7$\\
TMA~\cite{TMA} & $54.2$ & $73.8$ & $83.1$ & $90.2$\\
NSFT~\cite{NSFT} & $40.9$ & $64.7$ & $73.2$ & $81.0$\\
DML$^2$V~\cite{dml2v/SunWL17} &\textcolor{blue}{\bf64.5}&\textcolor{blue}{\bf85.7}&\textcolor{blue}{\bf92.1}& \textcolor{blue}{\bf96.0}\\
CSL~\cite{iccv15_correspondence} &  $44.4$ & $71.6$ & $82.2$ & $ 89.8$\\
\hline\hline
REGCT$_{best}(_{w/p})$ &  $\textcolor{red}{\bf 70.9}$ & \textcolor{red}{\bf 89.1} & \textcolor{red}{\bf 93.5} & \textcolor{red}{\bf 96.5}\\
\hline
\end{tabular}
\end{center}
%\vspace{-3ex}
\end{table}
On the 3DPES dataset, we compare the REGCT method with state-of-the-arts including LFDA~\cite{lfda}, ME~\cite{learn_to_rank}, kLFDA~\cite{kLFDA14}, PCCA~\cite{pcca}, rPCCA~\cite{kLFDA14}, SCSP~\cite{sim_spatial_constraints}, WARCA~\cite{WARCA}, DGD~\cite{DGD}, Spindle Net~\cite{Spindle_Net} and CSL~\cite{iccv15_correspondence}. As shown in Table~\ref{tab:3dpes_Dataset.}, the proposed algorithm significantly outperforms the state-of-the-art algorithms, and even deep learning based algorithm~\cite{Spindle_Net}. Note that the images in this dataset are automatic detection results from videos captured under eight cameras, bringing serious pose variations, illumination changes and scale variations. With the help of the learned correspondences templates, our REGCT model is robust against these issues.
\begin{table}[!tbp]
% \captionsetup{font={footnotesize}}
\caption{Comparison of top $r$ matching rate using CMC (\%) on 3DPES dataset. The best result is marked in red font, and the second best in blue.
\label{tab:3dpes_Dataset.}}
\begin{center}
\begin{tabular}{@{}C{3cm}@{}| @{}C{1.2cm}@{}@{}C{1.2cm}@{}@{}C{1.2cm}@{}@{}C{1.2cm}@{}}
\hline
Methods  & r=1 & r=5 & r=10 & r=20 \\
\hline\hline
LFDA~\cite{lfda} &  $45.5$ & $69.2$ & $-$ & $86.1$\\
ME~\cite{learn_to_rank} & $53.3$ & $76.8$ & $-$ & $92.8$\\
kLFDA~\cite{kLFDA14} &  $54.0$ & $77.7$ & $85.9$ & $92.4$\\
PCCA~\cite{pcca} &  $41.6$ & $70.5$ & $81.3$ & $90.4$\\
rPCCA~\cite{kLFDA14} &  $47.3$ & $75.0$ & $84.5$ & $91.9$\\
SCSP~\cite{sim_spatial_constraints} &  $57.3$ & $79.0$ & $-$ & $91.5$\\
WARCA~\cite{WARCA} &  $51.9$ & $75.6$ & $-$ & $-$\\
DGD~\cite{DGD} &  $56.0$ & $-$ & $-$ & $-$\\
Spindle-Net~\cite{Spindle_Net} &  $\textcolor{blue}{\bf62.1}$ & $\textcolor{blue}{\bf83.4}$ & $\textcolor{blue}{\bf90.5}$ & $\textcolor{blue}{\bf95.7}$\\
CSL~\cite{iccv15_correspondence} &  $57.9$ & $81.1$ & $89.5$ & $93.7$\\
\hline\hline
REGCT$_{best}(_{w/p})$ &  $\textcolor{red}{\bf 78.0}$ & $\textcolor{red}{\bf 92.4}$ & $\textcolor{red}{\bf 96.5}$ & $\textcolor{red}{\bf 99.1}$\\
\hline
\end{tabular}
\end{center}
%\vspace{-3ex}
\end{table}

On the CUHK01 dataset, we compare with Semantic~\cite{Semantic15},  kLFDA~\cite{kLFDA14}, IDLA~\cite{AhmedJM15}, DeepRanking~\cite{deep_ranking}, ME~\cite{learn_to_rank}, GOG~\cite{GOG}, SalMatch~\cite{salicency}, CSBT~\cite{CSBT}, TCP~\cite{DEEP_MULTI_CHANNEL} and DML$^2$V~\cite{dml2v/SunWL17}. The detailed comparison results are presented in Table~\ref{tab:cuhk_Dataset.}. As shown in Table~\ref{tab:cuhk_Dataset.}, the proposed graph matching and correspondence transfer framework can achieve favorable results on this medium-sized dataset. More specifically, the  proposed REGCT algorithm obtains best rank 1 recognition rate (a 14.3\% performance gain over TCP~\cite{DEEP_MULTI_CHANNEL}, a part based deep learning algorithm). And the better performance compared with other algorithms also demonstrate the superiority of the REGCT model.
\begin{table}[!tbp]
% \captionsetup{font={footnotesize}}
\caption{Comparison of top $r$ matching rate using CMC (\%) on CUHK01 dataset. The best result is marked in red font, and the second best in blue.
\label{tab:cuhk_Dataset.}}
\begin{center}
\begin{tabular}{@{}C{3cm}@{}| @{}C{1.2cm}@{}@{}C{1.2cm}@{}@{}C{1.2cm}@{}@{}C{1.2cm}@{}}
\hline
Methods  & r=1 & r=5 & r=10 & r=20 \\
\hline\hline
Semantic~\cite{Semantic15} &$32.7$&$51.2$ &$-$&$76.3$ \\
kLFDA~\cite{kLFDA14} &  $32.8$ & $59.0$ & $69.6$ & $-$\\
IDLA~\cite{AhmedJM15} &  $47.5$ & $71.5$ & $80.0$ & $-$\\
DeepRanking~\cite{deep_ranking} &  $50.4$ & $75.9$ & $84.1$ & $-$\\
ME~\cite{learn_to_rank} &  $53.4$ & $76.3$ & $84.4$ & $-$\\
GOG~\cite{GOG} &  $57.8$ & $79.1$ & $86.2$ & $-$\\
SalMatch~\cite{salicency} &$28.5$&$46.0$& $-$&$67.3$\\
CSBT~\cite{CSBT} &  $51.2$ & $76.3$ & $-$ & $91.8$\\
TCP~\cite{DEEP_MULTI_CHANNEL} &  $53.7$ & $84.3$ & $91.0$ & $\textcolor{red}{\bf96.3}$\\
DML$^2$V~\cite{dml2v/SunWL17} &$\textcolor{blue}{\bf65.0}$&$\textcolor{blue}{\bf85.6}$&$\textcolor{blue}{\bf91.1}$& $\textcolor{blue}{\bf95.1}$\\
\hline\hline
REGCT$_{best}(_{w/p})$ &  $\textcolor{red}{\bf 68.0}$ & $ \textcolor{red}{\bf86.9}$ & $ \textcolor{red}{\bf91.2}$ & $94.9$\\
\hline
\end{tabular}
\end{center}
\vspace{-3ex}
\end{table}

\subsection{Analysis on Typical Failure Cases}
\begin{figure}[!t]
  \centering
  \includegraphics[width=\linewidth]{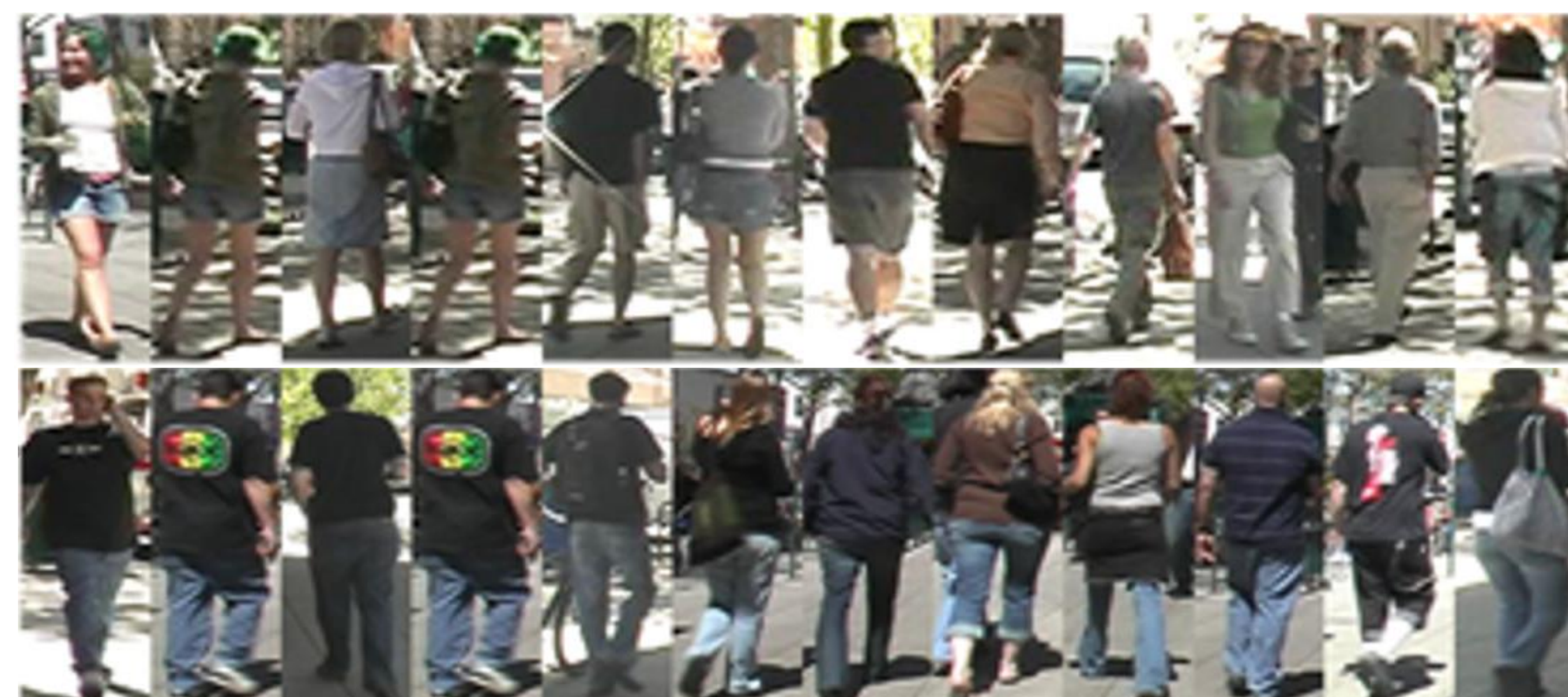}
%  \captionsetup{font={footnotesize}}
  \caption{Typical failure cases. The first image in each row is the probe image, the second one is the correctly matched gallery image, followed by the ranking list obtained by REGCT.   }
% \vspace{-2ex}
\label{fig:failure_cases}
\end{figure}

We record the typical failure cases to explore the limitations of the proposed REGCT algorithm. As shown in Fig.~\ref{fig:failure_cases}, when severe self-occlusion occurs, the appearances of the same person may be dramatically different across different camera views, rendering it difficult for REGCT to establish enough local correspondences between matched image pairs. Even though, the proposed REGCT algorithm can rank visually similar images with the probe ahead of others, which is valuable for further manual verification.

% or


%\appendices

%\section{}
\section{Conclusion}
\label{sec:conclusion}
This paper proposes a robust and efficient graph correspondence transfer (REGCT) approach to explicitly address the spatial misalignment issue in Re-ID. The framework of \emph{off-line} patch-wise correspondence learning and \emph{on-line} correspondence transfer helps to flexibly establish robust and accurate patch-level matching patterns for each test pair. The proposed template ensemble strategy is demonstrated to improve the efficiency as well as notably boost the recognition performance compared to the baseline GCT. The proposed pose context descriptor further benefits the REGCT model via more accurate templates selection, leading to more robust patch-wise correspondence transfer in the testing phase. Extensive experiments on five challenging datasets demonstrate the effectiveness of the REGCT model.

 %GCT model aims to learn a set of patch-wise correspondence templates from positive image pairs in the training set, and then transfer these correspondences to test image pairs with similar pose-pair configurations for distance computation. Owing to the part-based strategy as well as the incorporation of the body context information, the GCT model is capable of dealing with the problem of spatial misalignment caused by large variations in viewpoints and human poses. Extensive experiments on five challenging datasets demonstrate the effectiveness of the GCT model.\\

%% use section* for acknowledgment
\section*{Acknowledgment}
This work was supported by the National Natural Science Foundation of China (NSFC) (Grant No. 61671289, 61221001, 61528204, 61771303, 61521062 and 61571261), STCSM (18DZ2270700) and by US National Science Foundation (NSF) (Grant No.1618398 and 1350521).

\ifCLASSOPTIONcaptionsoff
  \newpage
\fi



\bibliographystyle{IEEEtran}
\bibliography{reference}
%\begin{thebibliography}{1}


%\end{thebibliography}

%\vspace{-2.5 cm}
\begin{IEEEbiography}[{\includegraphics[width=1in,height=1.25in,clip,keepaspectratio]{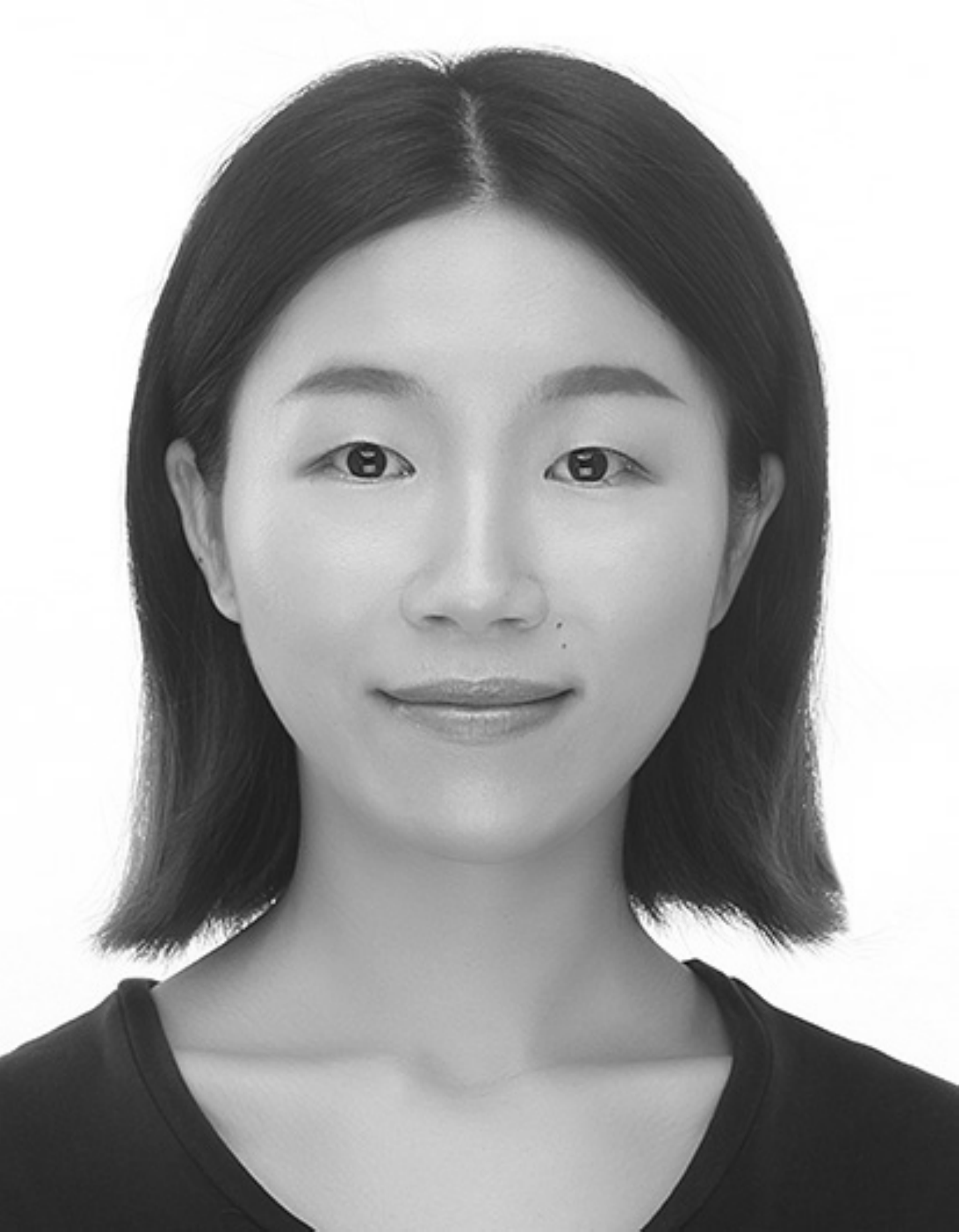}}]{Qin Zhou}
received her B.S. degree in information engineering from
Xi'an Jiao Tong University, Xi'an, China, in 2013.
    She is currently a Ph.D. student at the Department of Electrical Engineering, Shanghai Jiao Tong University,
Shanghai, China. Her current research interests
include computer vision, machine learning, convex optimization and person re-identification. She is now a visiting student in Professor Haibin Ling's lab at Temple University.
\end{IEEEbiography}
\vspace{-0.3 cm}
\begin{IEEEbiography}[{\includegraphics[width=1in,height=1.25in,clip,keepaspectratio]{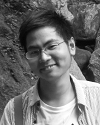}}]{Heng Fan}
 received his B.E. degree in College of Science, Huazhong Agricultural University (HZAU), Wuhan, China, in 2013. He is currently a Ph.D. student in the Department of Computer and Information Science, Temple University, Philadelphia, USA. His research interests include computer vision, pattern recognition and machine learning.
\end{IEEEbiography}
\vspace{-0.3 cm}
\begin{IEEEbiography}[{\includegraphics[width=1in,height=1.25in,clip,keepaspectratio]{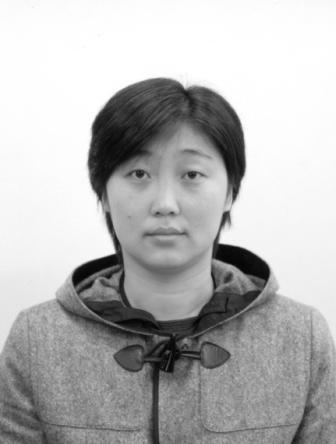}}]{Hua Yang} received her Ph.D. degree in communication and information from Shanghai Jiaotong University, in 2004, and both the B.S. and M.S. degrees in communication and information from Haerbin Engineering University, China in 1998 and 2001, respectively. She is currently an associate professor in the Department of Electronic Engineering, Shanghai Jiaotong University, China. Her current research interests include video coding and networking, computer vison, and smart video surveillance.
\end{IEEEbiography}
\vspace{-0.3 cm}
\begin{IEEEbiography}[{\includegraphics[width=1in,height=1.25in,clip,keepaspectratio]{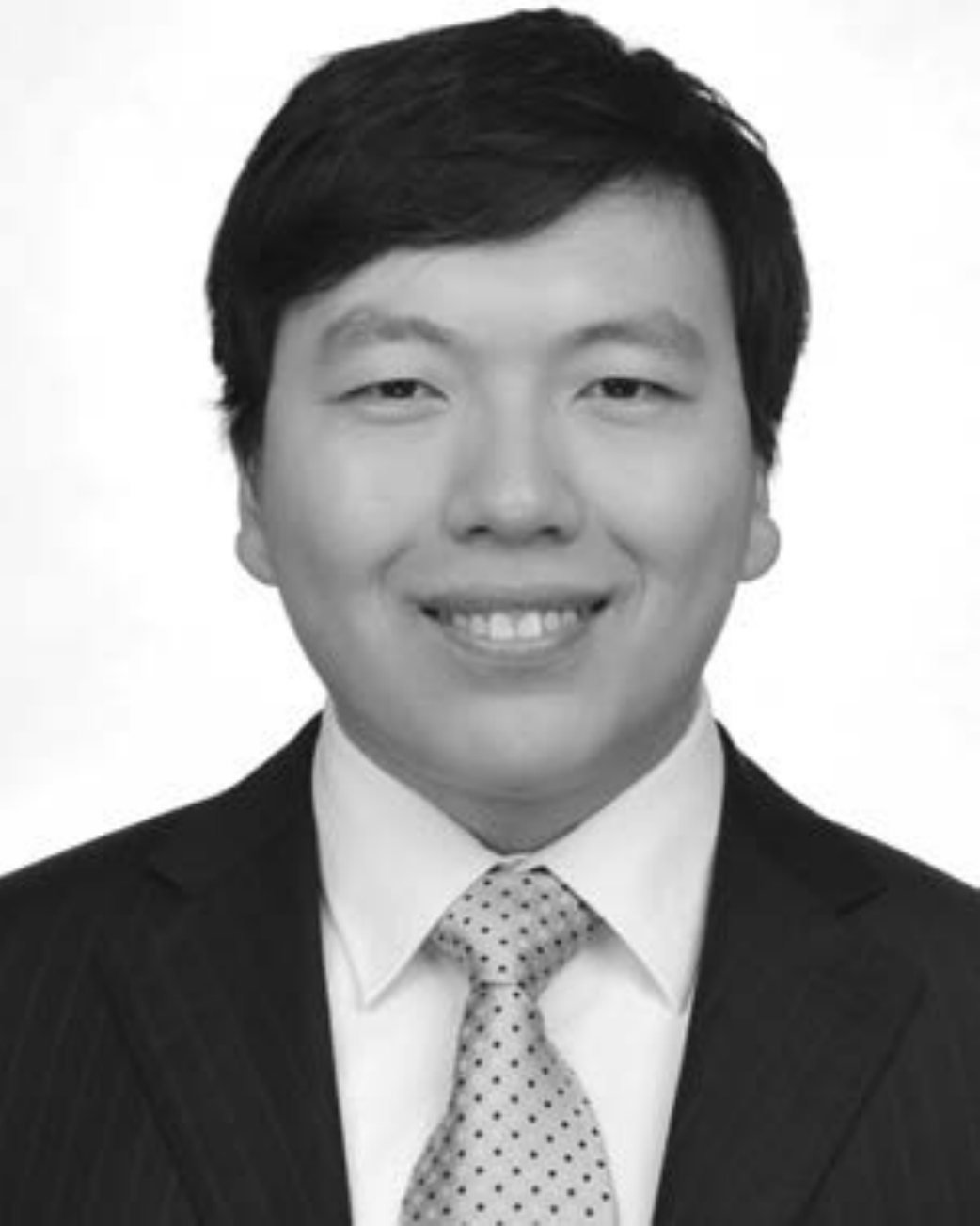}}]{Hang Su},
received his Ph. D Degree in Electronic Engineering from Shanghai Jiaotong University in 2014. He is currently working as a PostDoc in department of Computer Science, Tsinghua University. His current research interests  include computer vision and large-scale machine learning. Mr. Hang Su has also served as TPC memember in several conferences including IJCAI, AAAI and UAI, and contributed as a reviewer for TPAMI, CVPR, NIPS, ICML, etc.
\end{IEEEbiography}
\vspace{-0.3 cm}
\begin{IEEEbiography}[{\includegraphics[width=1in,height=1.25in,clip,keepaspectratio]{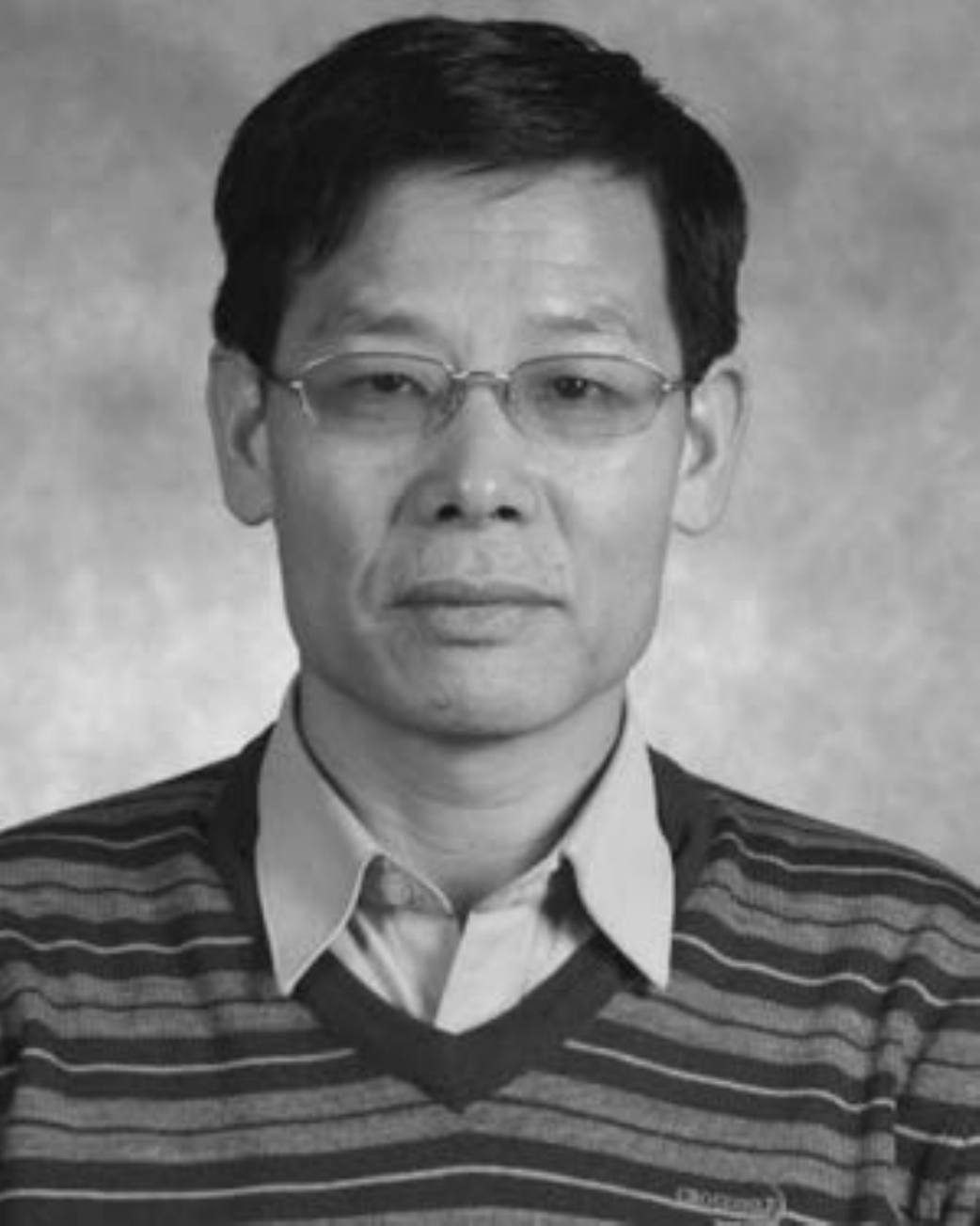}}]{Shibao Zheng}
received his B.S. degree in communication engineering from Xidian
University, Xi'an and M.S.degree in  the signal and information processing from the 54th institute of CETC, Shijiazhuang, China, in 1983 and 1986, respectively.
He is currently a professor of electronic engineering department and vice director
of Elderly Health Information and Technology
Institute, Shanghai Jiao Tong University (SJTU),
Shanghai, China. And he is also a professor
committee member of Shanghai Key Laboratory
of Digital Media Processing and Transmission, and
a Consultant Expert of ministry of public security in video surveillance field. His current research interests include urban video surveillance system, intelligent video analysis,  and elderly health technology, etc.
\vspace{-0.3 cm}
\end{IEEEbiography}

\begin{IEEEbiography}[{\includegraphics[width=1in,height=1.25in,clip,keepaspectratio]{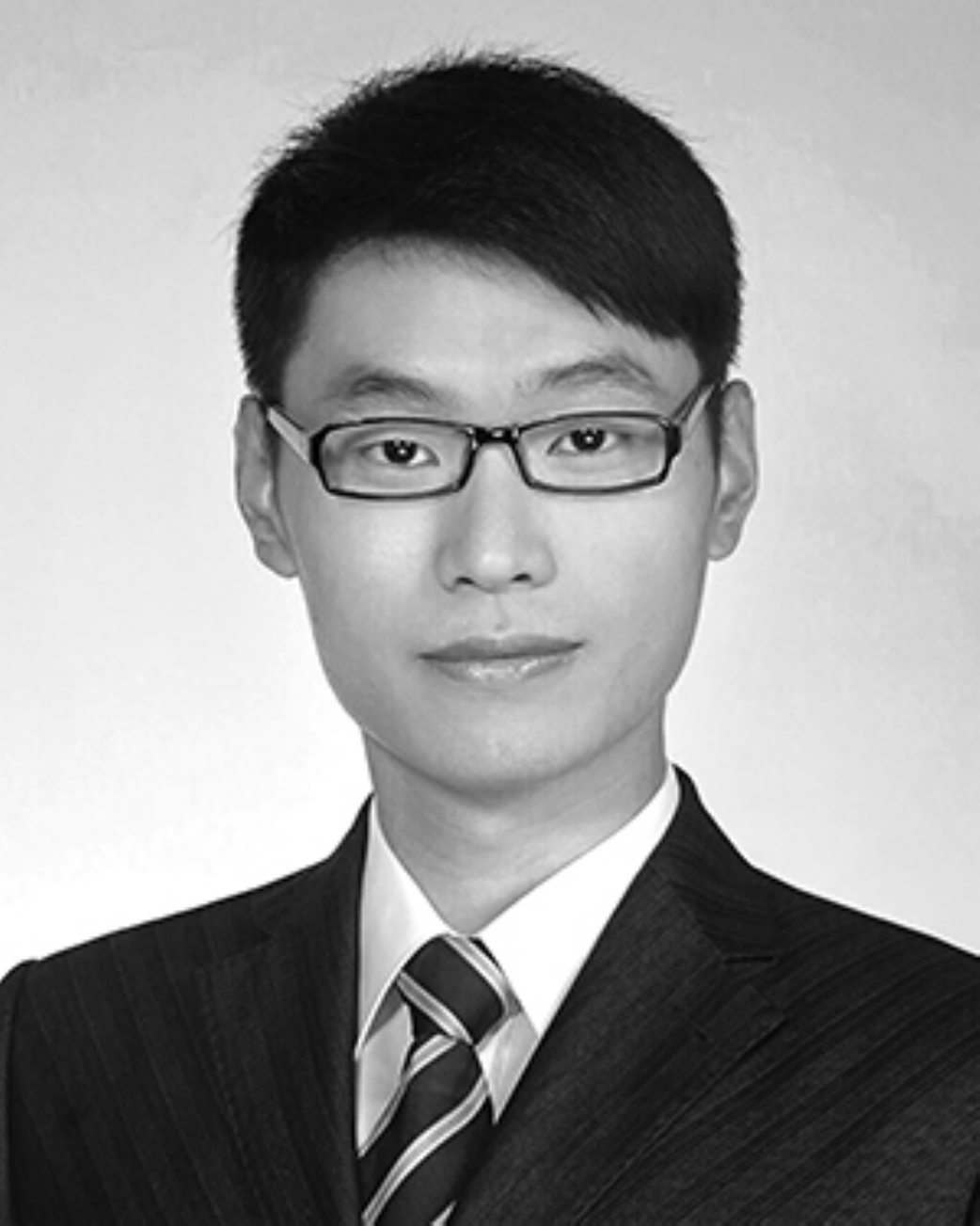}}]{Shuang Wu}
received the BS degree in electronic information engineering from
the Southeast University (SEU), Nanjing, China, in 2010.
He is currently a Ph.D. student at the Department of Electrical Engineering, Shanghai Jiao Tong University,
Shanghai, China. His current research interests
include computer vision, machine learning and crowd behavior analysis. He is now a researcher at Youtu Lab, Tencent.
\end{IEEEbiography}
\vspace{-0.3 cm}
\begin{IEEEbiography}[{\includegraphics[width=1in,height=1.25in,clip,keepaspectratio]{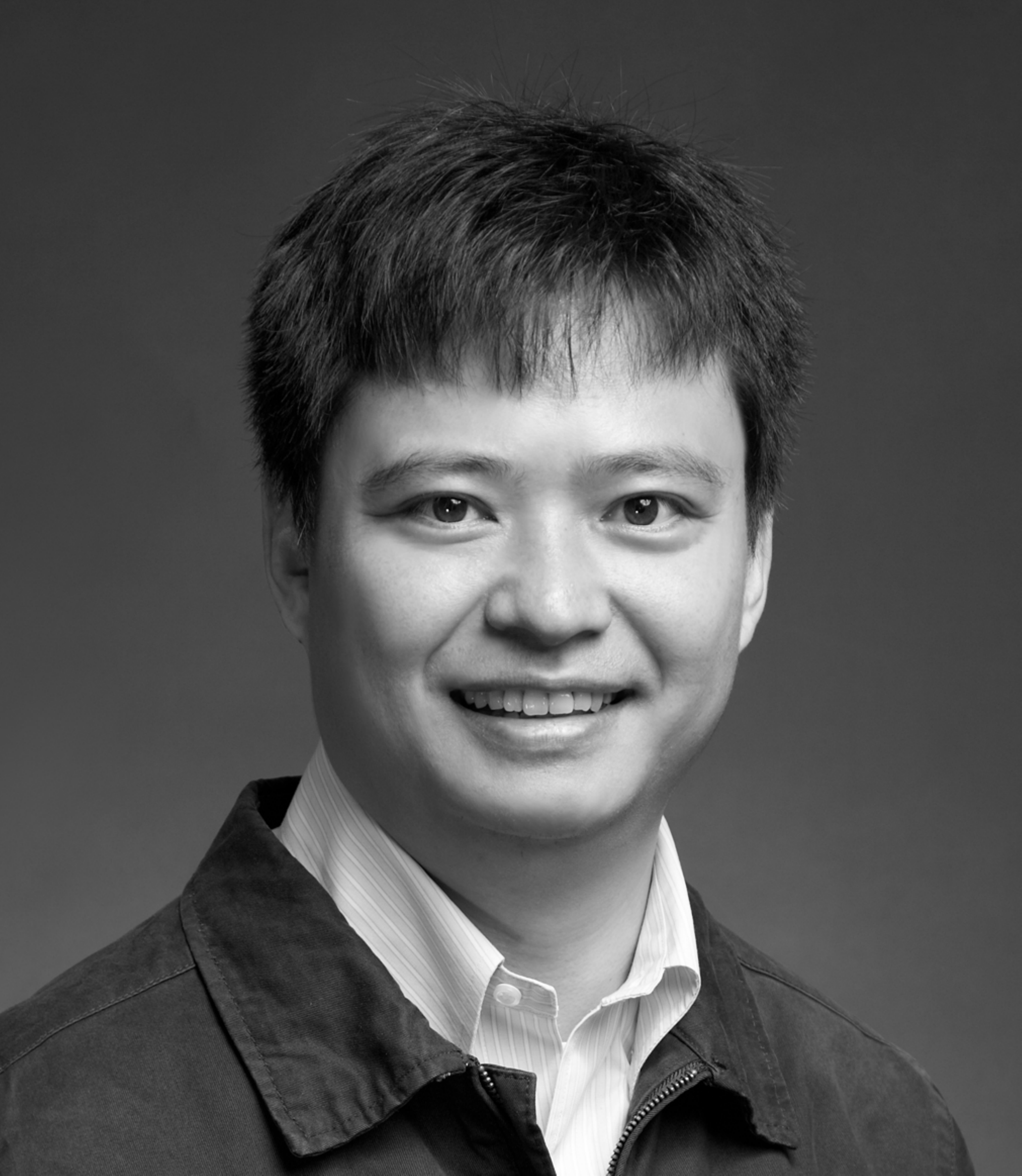}}]{Haibin Ling}
received the BS and MS degrees from Peking University, China, in 1997 and 2000, respectively, and the PhD degree from the University of Maryland College Park in 2006. From 2000 to 2001, he was an assistant researcher at Microsoft Research Asia. From 2006 to 2007, he worked as a postdoctoral scientist at the University of California Los Angeles. After that, he joined Siemens Corporate Research as a research scientist. Since fall 2008, he has been with Temple University where he is now an Associate Professor. Ling's research interests include computer vision, augmented reality, medical image analysis, and human computer interaction. He received the Best Student Paper Award at the ACM Symposium on User Interface Software and Technology (UIST) in 2003, and the NSF CAREER Award in 2014. He serves as associate editors for {\it IEEE Transactions on Pattern Analysis and Machine Intelligence}, {\it Pattern Recognition}, and {\it Computer Vision and Image Understanding}, and has served as area chairs for CVPR 2014 and CVPR 2016.
\end{IEEEbiography}

%\newpage


%\vfill


\end{document}